%% file: main.tex
\pdfoutput=1

\documentclass[10pt,twocolumn,letterpaper]{article}

\usepackage[pagenumbers]{cvpr} %

\input{sections/template_support/preamble}
\definecolor{cvprblue}{rgb}{0.21,0.49,0.74}
\usepackage[pagebackref,breaklinks,colorlinks,allcolors=cvprblue]{hyperref}
\usepackage{tcolorbox}
\usepackage[table]{xcolor}
\providecolor{oursred}{RGB}{255,236,236}
\newif\ifhrfinal
\hrfinaltrue

\usepackage{algorithm,algpseudocode}
\newlength{\algtrimlen}
\newcommand{\algtrim}[1]{\global\algtrimlen=#1\relax}
\makeatletter
\renewcommand\fs@ruled{\def\@fs@cfont{\bfseries}\let\@fs@capt\floatc@ruled
  \def\@fs@pre{\hrule height.8pt depth0pt \kern2pt}%
  \def\@fs@post{\kern2pt\hrule\relax\kern-\algtrimlen\global\algtrimlen=\z@}%
  \def\@fs@mid{\kern2pt\hrule\kern2pt}%
  \expandafter\let\expandafter\@fs@iftopcapt\csname iftrue\endcsname}
\makeatother
\usepackage{booktabs,multirow}
\usepackage{pifont}
\newcommand{\tabyes}{\textcolor{green!55!black}{\ding{51}}}
\newcommand{\tabno}{\textcolor{black!30}{\ding{55}}}
\usepackage{cuted}
\usepackage{caption}
\usepackage{nicematrix}
\usepackage{tikz}
\usepackage{amssymb,tabularx}

\algrenewcommand{\algorithmicindent}{0.8em}          
\makeatletter\renewcommand{\ALG@beginalgorithmic}{\small}\makeatother

\makeatletter
\renewcommand{\paragraph}{\@startsection{paragraph}{4}{\z@}
  {1.3ex \@plus .5ex \@minus .2ex}   
  {-1em}%
  {\normalfont\normalsize\bfseries}}
\makeatother

\newtcolorbox{insight}[1]{%
  colback=gray!10, colframe=black, arc=0pt, outer arc=0pt,
  toprule=0.6pt, bottomrule=0.6pt, leftrule=0pt, rightrule=0pt,
  boxsep=0pt, left=3pt, right=3pt, top=2.5pt, bottom=2.5pt,
  before skip=3pt plus 2pt, after skip=5pt plus 2pt,
  before upper={\textbf{#1.}\ }}
\def\confName{CVPR}
\def\confYear{2027}

\title{PoCoAttn: Position--Content Collaborative Sparse Attention for\\Fast and High-Fidelity Video Generation}

\author{%
Jianwen Wu, Meiling Wang, Junkang Zhou, Yipeng Shen, Haoran You$^{\dagger}$\\[0.3em]
Purdue University\\[0.3em]
{\normalsize Project page: \url{https://ecoai-group.github.io/pocoattn}}
}

\newcommand{\authorfootnote}[2]{\begin{NoHyper}\renewcommand{\thefootnote}{\ensuremath{#1}}\footnotetext{#2}\end{NoHyper}}

\begin{document}
\maketitle
\authorfootnote{\dagger}{Corresponding author. Email: \href{mailto:haoran@purdue.edu}{\texttt{haoran@purdue.edu}}}
\input{sections/00_abstract}

\input{sections/01_introduction_v2}

\input{sections/02_related_work}

\input{sections/03_preliminaries}

\input{sections/04_methodology_v2}

\input{sections/05_experiments_v2}

\input{sections/06_conclusion}

\input{sections/07_appendix}

\end{document}

%% file: sections/00_abstract.tex
\maketitle
\begin{strip}
\vspace{-18mm}
  \centering
  \includegraphics[width=\textwidth]{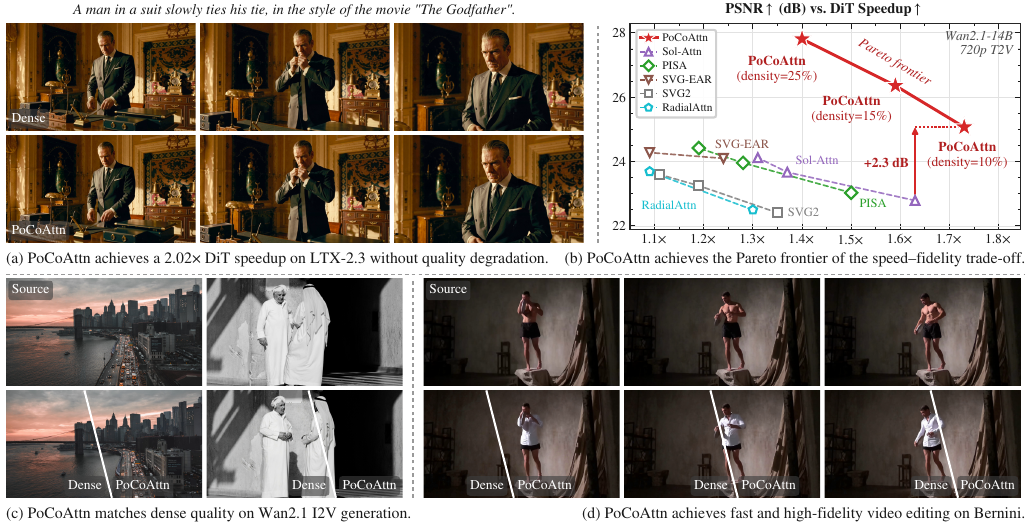}
  \captionsetup{type=figure,hypcap=false}
  \vspace{-1.5em}
  \captionof{figure}{Generated samples of PoCoAttn compared with dense attention, and speed--fidelity trade-off against sparse attention baselines.}
  \label{fig:teaser}
  \vspace{-0.5em}
\end{strip}

\begin{abstract}

\vspace{-0.5em}
Video diffusion transformers rely on 3D full attention over tens of thousands of spatiotemporal tokens, and its quadratic cost dominates inference latency.
Sparse attention reduces this cost, but existing methods face a dilemma: static methods apply predefined masks and cannot capture input-dependent attention, whereas dynamic methods reconstruct the sparse mask repeatedly, paying online to rediscover positional structure that is largely stable across inputs.
To this end, we propose \textbf{PoCoAttn}, a training-free sparse attention framework that separates input-stable from input-dependent attention and resolves the former offline and the latter online. PoCoAttn features:
(1) A relative-offset analysis of 3D query--key pairs, which shows that video attention decomposes into an input-stable positional prior that carries most of the attention mass and a small content-dependent residual.
(2) A position--content collaborative sparse attention mechanism, which retrieves positional keys from offline-calibrated tables, selects content keys online with a head-adaptive ratio between the two sources, and compensates for the discarded keys to approximate full attention.
(3) A step-adaptive budget allocation algorithm, which models how early-step sparse attention errors are amplified along the denoising trajectory and derives the optimal sparsity budget per step.
Extensive experiments across four video diffusion models and three tasks show that PoCoAttn consistently achieves a better speed--fidelity Pareto frontier than state-of-the-art (SOTA) methods, improving PSNR by 1.9 to 5.0 dB over baselines and delivering up to 4.1$\times$ attention and 2.0$\times$ denoising speedups over dense FlashAttention-3 attention. 

\end{abstract}

%% file: sections/01_introduction_v2.tex
\vspace{-1em}
\section{Introduction}
\label{sec:intro}
Video diffusion transformers~\cite{wan,hunyuanvideo,ltx2} have become the leading approach to high-fidelity video generation, yet their inference cost limits practical deployment.
They rely on 3D full attention over tens of thousands of spatiotemporal tokens, whose quadratic cost dominates inference latency at high resolution and long duration~\cite{svg,sta}.
For instance, HunyuanVideo requires approximately 16 minutes to generate a 129-frame 720p video comprising 119K tokens on an NVIDIA H100 GPU, with attention accounting for over 80\% of denoising latency even using FlashAttention-3~\cite{fa3}.
Video attention is also highly sparse, concentrating most of its mass on a small subset of keys~\cite{svg,svg2}, which makes training-free sparse attention a natural remedy: compute only the important query--key interactions and skip the rest.
Its effectiveness, however, hinges on identifying those keys accurately and cheaply, since missed keys introduce errors that propagate through the denoising trajectory, while heavy online selection can negate the savings from sparsification.

Existing methods face a dilemma between adaptivity and selection overhead.
Static methods apply predefined masks, either hand-crafted~\cite{svg,sta,radial} or calibrated offline from the input-stable part of attention~\cite{calibatt,scalingattn}.
They incur little selection overhead but cannot capture input-dependent attention because the entire mask is fixed, so fidelity degrades quickly at higher sparsity.
Dynamic methods instead construct sparse masks online through proxy-based scoring~\cite{xattention,spargeattn,pisa,sol}, token clustering~\cite{svoo,svg2,svgear}, or exact attention mass~\cite{mcsparse}.
They adapt to content, but reconstruct the mask repeatedly during inference, paying online to rediscover positional structure that is largely stable across inputs.
Neither family separates the input-stable part of attention from the input-dependent part, so neither can determine the former offline while selecting the latter online.

In this work, we aim to resolve this dilemma by proposing \textbf{PoCoAttn}, the first training-free sparse attention framework that separates input-stable from input-dependent attention and resolves the former offline and the latter online. PoCoAttn answers three questions.
\textit{\textbf{First}, which components of attention are input-stable, and which are input-dependent and thus need online selection?}
We analyze attention over relative 3D query--key offsets, a view that subsumes hand-crafted spatial and temporal patterns as particular offset sets. This analysis shows that each head's attention decomposes into an input-stable \emph{positional prior}, the mean attention at each offset, and a \emph{content residual}. The prior is highly consistent across prompts and alone retains most of the attention mass even under a small key budget, so it can be calibrated once offline, leaving only the content residual to online selection.
\textit{\textbf{Second}, how can the positional prior and the content residual be derived and what ratio between them should be used?}
Positional keys are retrieved from offline-calibrated tables, and content keys are selected online.
We find that the fraction of attention the positional prior captures varies widely across heads yet remains stable across inputs for each head, so the position--content ratio can be set per head offline.
The mass the prior misses further splits into a few content-specific keys that online selection recovers, and a diffuse tail over many discarded keys that should be compensated for rather than selected.
\textit{\textbf{Third}, how should a sparsity budget be distributed across denoising steps?}
Existing methods apply either uniform or heuristic schedules~\cite{svg2,pisa,sol,dfsattn,rapid,pasa} without modeling how sparse attention errors propagate. Unlike token pruning~\cite{diffratio}, we observe that early-step errors are both larger and far more amplified along the denoising trajectory than later-step errors, so the budget should be front-loaded based on a measured amplification profile rather than spread evenly.
Our contributions are summarized as follows:

\begin{itemize}
    \item We propose \textbf{PoCoAttn}, a training-free sparse attention framework built on a relative-offset analysis of video attention that generalizes hand-crafted sparse patterns into a unified, more expressive search space, and decomposes attention into input-stable and input-dependent parts.

    \item We design a position--content collaborative sparse attention mechanism that retrieves the positional prior from offline-calibrated tables, selects the content residual online with a head-adaptive ratio, and compensates for discarded keys with pooled virtual keys in the softmax.
    
    \item We develop a step-wise sparsity allocation algorithm that models the step-dependent amplification of sparse attention errors, measured once per model, and determines the optimal budget per step given an overall sparsity ratio.
\end{itemize}

Extensive experiments on four video diffusion models and three tasks show that PoCoAttn consistently achieves a better speed--fidelity Pareto frontier than SOTA methods, \textbf{improving PSNR by 1.9 to 5.0~dB} over baselines at lower density and delivering \textbf{up to 4.1$\times$ attention and 2.0$\times$ denoising speedups} over dense FlashAttention-3 attention.

%% file: sections/02_related_work.tex
\begin{figure*}[!t]
\centering
\includegraphics[width=\textwidth]{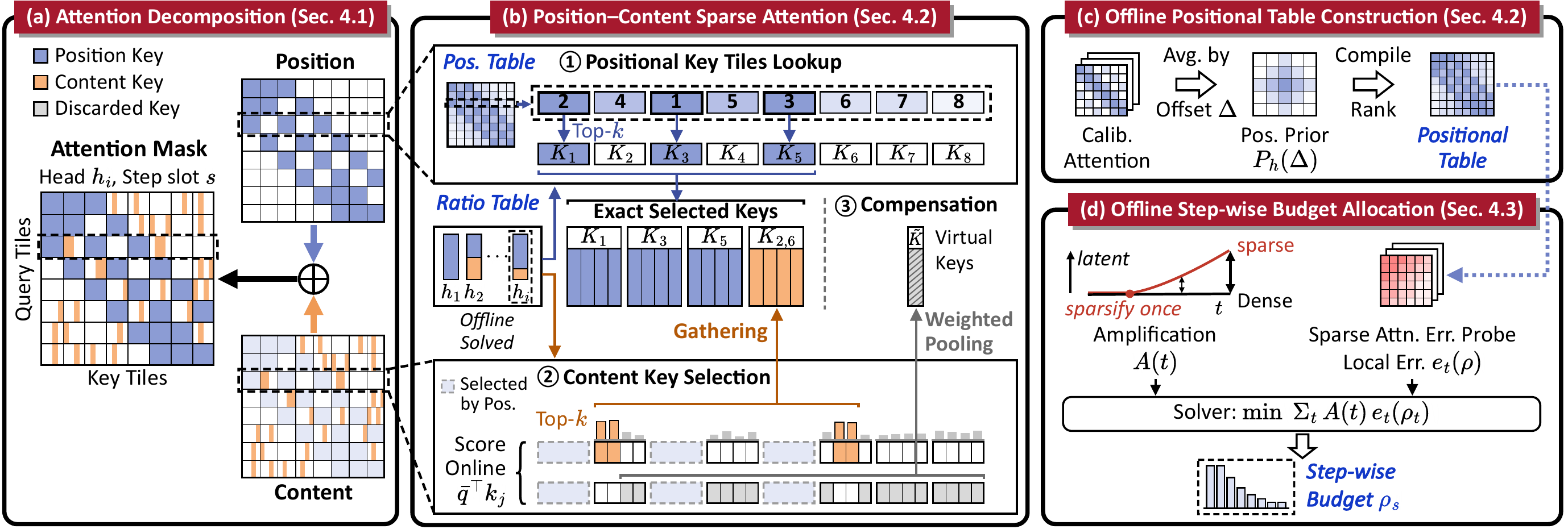}
\vspace{-1.5em}
\caption{\textbf{Overview of PoCoAttn.} 
(a) PoCoAttn decomposes attention into an input-stable \textit{positional} part and an input-dependent \textit{content} part.
(b) At inference, each query tile looks up its positional key tiles in the positional table, selects content keys by pooled-query scores, and merges the discarded keys into virtual keys by weighted pooling; the per-head position--content ratio is solved offline.
(c) The positional table is compiled offline from the per-head positional prior $P_h(\Delta)$, obtained by averaging calibration attention over each relative offset $\Delta$.
(d) The step-wise budget $\rho_s$ is solved offline by minimizing the local error $e_t(\rho_t)$ weighted by its amplification $A(t)$.}
\label{fig:overview}
\vspace{-1.em}
\end{figure*}

\section{Related Work}

\paragraph{Sparse Attention for Video Generation.}
\textit{Static methods} derive their masks from a pattern family fixed before inference: hand-crafted templates such as sliding 3D windows~\cite{sta}, distance-decaying bands~\cite{radial}, spatial--temporal patterns~\cite{svg}, and diagonal or stripe menus~\cite{sparsevdit,moddit}, or masks calibrated offline from full attention maps~\cite{calibatt,scalingattn,psa}.
Some choose or combine patterns within the family at run time~\cite{svg,moddit}, but remain constrained by its predefined structure, limiting their ability to capture content-dependent interactions.
\textit{Dynamic methods} construct input-dependent masks, via cheap block proxies~\cite{spargeattn,xattention,pisa,sol}, token clustering~\cite{svg2,svgear,svoo}, exact attention mass~\cite{mcsparse}, or learned selection~\cite{vsa}, and some compensate for the discarded keys~\cite{pisa,sol,svgear,mcsparse}. Although adaptive to content, these methods search across the full key space at each attention call or designated anchor step.
PoCoAttn takes most keys from an offline, head-specific positional prior that generalizes static patterns, searches online only for a small share of content keys, and compensates for the rest.

\paragraph{Adaptive Sparsity Budget Allocation.}
Most methods apply a fixed sparsity budget across denoising steps~\cite{pisa,sol,svg2}, despite the unequal impact of different steps on final video quality.
Luo~\etal~\cite{svoo} and HEART~\cite{heart} adapt sparsity across layers or heads but keep the budget fixed across steps.
For step-wise budget adaptation, Zhang~\etal~\cite{pasa} adjust density based on velocity changes between consecutive steps, while RAPID~\cite{rapid} switches to a sparser mask at a step chosen by attention recall.
However, these methods rely on local signals or hand-crafted schedules without accounting for how errors propagate to the final video.
We instead model this propagation, profile step-wise error amplification once offline, and solve for the budget allocation that minimizes amplification-weighted attention error.

%% file: sections/03_preliminaries.tex
\section{Preliminaries}
\label{prelim}

\paragraph{Block-Sparse Attention.}
Given queries, keys, and values $\mathbf Q,\mathbf K,\mathbf V \in \mathbb R^{N\times d}$, dense attention computes $\mathbf O=\operatorname{Softmax}(\mathbf Q\mathbf K^\top/\sqrt d)\mathbf V$, incurring quadratic complexity in the sequence length $N$. Block-sparse attention (BSA) reduces this cost by partitioning queries and keys into tiles and computing attention only for selected tile pairs:
$\widehat{\mathbf O}=\operatorname{Softmax}(\mathbf Q\mathbf K^\top/\sqrt d+\mathbf M)\mathbf V$,
where $M_{ij}=0$ for retained interactions and $-\infty$ otherwise. This tile-wise structure enables efficient %
execution by skipping unselected tiles.

\paragraph{3D Rotary Position Embeddings.}
Video diffusion transformers use 3D rotary position embeddings (RoPE) to encode spatiotemporal positions into queries and keys. For tokens $i$ and $j$ at positions $\mathbf p_i=(t_i,y_i,x_i)$ and $\mathbf p_j=(t_j,y_j,x_j)$, the attention logit is $s_{ij}=\mathbf q_i^\top R(\Delta_{ij})\mathbf k_j/\sqrt d$, where $\mathbf q_i$ and $\mathbf k_j$ are the unrotated query and key, $\Delta_{ij}=\mathbf p_j-\mathbf p_i=(\Delta t,\Delta y,\Delta x)$ is their relative offset, and $R(\Delta_{ij})$ denotes the frequency-dependent rotation.

%% file: sections/04_methodology_v2.tex
\section{The Proposed PoCoAttn Framework}

\textbf{Overview.}
PoCoAttn selects each query's keys from two sources: positional keys retrieved from offline tables and content keys selected online.
In Sec.~\ref{sec:motivation}, it decomposes attention into an input-stable positional prior and an input-dependent content residual (Fig.~\ref{fig:overview}~(a)).
In Sec.~\ref{sec:poco}, it compiles the prior offline into positional tables (Fig.~\ref{fig:overview}~(c)) and at inference combines table lookup with online content-key selection and discarded-key compensation (Fig.~\ref{fig:overview}~(b)).
In Sec.~\ref{sec:step}, it allocates the sparsity budget across timesteps by minimizing the amplification-weighted error (Fig.~\ref{fig:overview}~(d)).

\begin{figure*}[t]
  \centering
  \includegraphics[width=\textwidth]{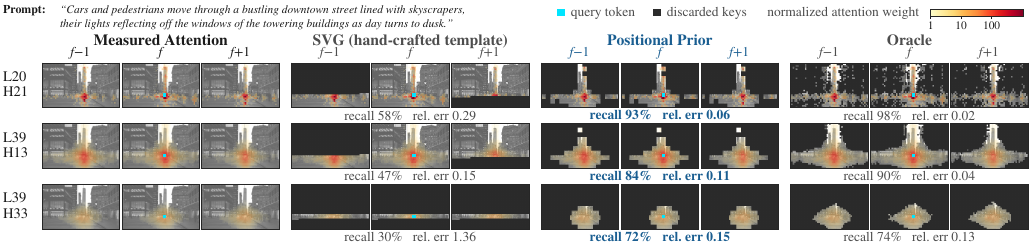}
  \vspace{-1.5em}
  \caption{\textbf{Relative position alone captures substantial attention mass.}
  Attention maps of three representative Wan2.1-14B heads at denoising step 10, for a query in latent frame $f{=}10$ and keys in frames $f{-}1$, $f$, $f{+}1$. Under the same 15\% key budget, the positional prior extracted from the same prompt captures head-specific patterns missed by hand-crafted templates and approaches the per-query oracle.
  Recall measures the retained attention mass; rel.\ err.\ measures the relative $\ell_2$ error of the sparse attention output against the dense one.}
  \label{fig:prior_vs_template}
  \vspace{-1.em}
\end{figure*}

\begin{figure}[t]
  \centering
  \includegraphics[width=\columnwidth]{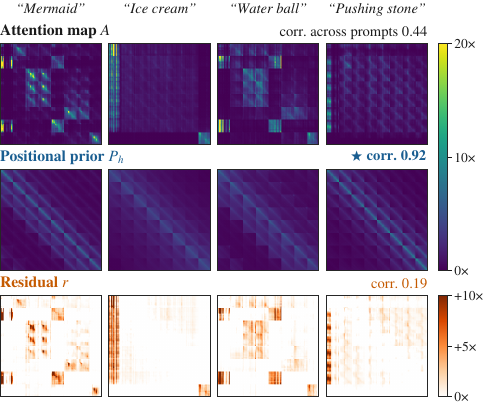}
  \vspace{-1.8em}
  \caption{\textbf{Attention decomposes into an input-stable positional prior and an input-dependent content residual.}
  Each column corresponds to a different prompt. The bottom row shows the residual attention that content adds on top of the positional prior.}
  \label{fig:attn_decomp}
  \vspace{-1.5em}
\end{figure}

\subsection{Attention Decomposition by Input Dependence}
\label{sec:motivation}
\paragraph{Relative-Offset Formulation.}
As discussed in Sec.~\ref{prelim}, 3D RoPE encodes positional dependence through $\Delta_{ij}$. We thus characterize query--key interactions by their relative offsets $\Delta\in\mathbb Z^3$. Under this view, existing hand-crafted offset-based patterns can be expressed as masks induced by an offset set $\mathcal D$:
$\mathcal M_{\mathcal D}=\{(i,j):\Delta_{ij}\in\mathcal D\}$.
For example, in SVG~\cite{svg}, the spatial pattern corresponds to $\mathcal D=\{\Delta:\Delta t=0\}$, while the temporal pattern corresponds to $\mathcal D=\{\Delta:\Delta x=\Delta y=0\}$.
Instead of restricting attention to predefined templates, this formulation provides a unified and expressive search space for discovering arbitrary head-specific offset patterns.
This view also separates what position explains in attention from what it does not.
For a head $h$, let $A_{ij}$ be the attention weight from query $i$ to key $j$. We define the \textbf{\emph{positional prior} $P_h$} as the average attention weight at each relative offset, and the \emph{residual} $r_{ij}$ as what remains:
\begin{equation}
\begin{aligned}
P_h(\Delta)&=\mathbb E\!\left[A_{ij}\mid\Delta_{ij}=\Delta\right],\\
A_{ij}&=\underbrace{P_h(\Delta_{ij})}_{\text{Positional prior}}+\underbrace{r_{ij}}_{\text{Residual}},
\end{aligned}
\label{eq:decomp}
\end{equation}
where the expectation is taken over query--key pairs and input samples. By construction, $P_h$ is the least-squares optimal approximation of the attention of head $h$ by a function of the relative offset alone, so the residual is zero-mean at each relative offset and carries the part of attention that no function of the relative offset can capture. A more detailed theoretical explanation is provided in App.~\ref{app:decomp}.

\paragraph{Corroborating Empirical Observations.}
To characterize the underlying input-stable structure of attention, we compare the two terms in Eq.~\eqref{eq:decomp} across inputs.
As shown in Fig.~\ref{fig:attn_decomp}, positional priors estimated from individual prompts remain highly consistent, with a mean pairwise correlation of $0.92$, while full attention maps and residuals vary substantially. Moreover, the positional prior captures a substantial fraction of the attention mass, providing a strong backbone for BSA. As indicated in Fig.~\ref{fig:prior_vs_template}, at a 15\% key budget, selecting keys by $P_h$ retains 72--93\% of the attention mass, approaching the per-query oracle and substantially outperforming hand-crafted templates~\cite{svg}. 

\begin{insight}{Insight 1}
Attention maps vary across prompts, but their positional priors vary far less and alone retain a substantial fraction of the attention mass at a small key budget.
Therefore, the prior can be calibrated once offline and reused, leaving only the content residual to online selection.
\end{insight}

\begin{figure}[t]
  \centering
  \includegraphics[width=\columnwidth]{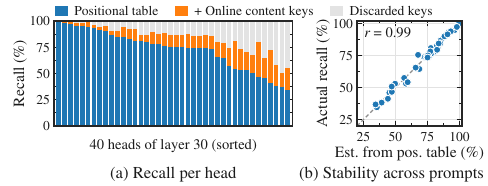}
  \vspace{-2em}
  \caption{\textbf{Positional recall is head-dependent but consistent across inputs.}
  (a) Recall of the 40 heads in layer 30 of Wan2.1-14B at a 25\% key budget: positional table alone, gain from moving half the budget to online content keys, and the mass left on discarded keys.
  (b) Per-head recall of the positional table: offline estimate (8 calibration prompts) versus actual on 8 unseen prompts.}
  \label{fig:prior_coverage}
  \vspace{-1.7em}
\end{figure}

Although the positional prior captures a substantial fraction of the attention mass, the remainder still depends on the input. We therefore examine how the retained mass differs across heads, using the table calibrated and compiled from the positional prior in Sec.~\ref{sec:poco}.
As shown in Fig.~\ref{fig:prior_coverage}~(a), the retained mass varies substantially across heads, ranging from 34.6\% to 99.7\% within a single layer at a 25\% key budget. The mass not captured by the positional prior comprises two components: \textit{The first} is input-dependent and concentrated on a few specific keys. It cannot be tabulated offline, but online selection based on query--key similarity recovers a large portion of it, with greater gains for heads with lower recall. \textit{The second} is a thin tail over the many discarded keys, negligible individually yet non-trivial in sum, and unrecoverable by selection within the budget. Despite this variation across heads, Fig.~\ref{fig:prior_coverage}~(b) indicates that for a given head, the mass retained by the positional table remains stable across prompts and closely follows its offline estimate along $y=x$, with a correlation of $0.99$.

\begin{insight}{Insight 2}
The attention mass captured by positional priors varies substantially across heads but remains consistent across inputs within each head, so each head's position--content ratio can be set offline.
\end{insight}

\begin{algorithm}[t]
\caption{Offline positional table construction} 
\algtrim{8pt} %
\label{alg:table}
\begin{algorithmic}[1]
\Require dense model; calibration prompts $\mathcal C$; key budget $\rho$; kernel tile size $G$; candidate box shapes $\mathcal B$
\For{each step slot, layer, head $h$}
  \State $P_h(\Delta) \gets \frac{1}{|\mathcal C|}
    \sum_{c\in\mathcal C} P_h^{(c)}(\Delta)$
    \Comment{\textcolor{red}{calibrate}}
  \State $b_h\gets\arg\max_{b\in\mathcal B}$ attn mass of $P_h$ kept at $\rho$ \Comment{\textcolor{red}{compile}}
  \State $\pi_h \gets$ pack $b_h$-boxes into contiguous $G$-token tiles
  \For{each query tile $a$}
    \State $m_h(a,g) \gets
      \sum_{i\in a,\,j\in g} P_h(\Delta_{ij})$
      \quad for each key tile $g$
    \State $\mathrm{order}_h(a) \gets
      \operatorname{argsort}_{g}^{\downarrow}\,m_h(a,g)$
  \EndFor
\EndFor
\Ensure positional table $\{\pi_h,\ \mathrm{order}_h\}$ per step slot, layer, head
\end{algorithmic}
\end{algorithm}

\subsection{Position--Content Sparse Attention}
\label{sec:poco}
\paragraph{Constructing the Positional Prior.}

Insight~1 suggests calibrating the positional prior offline and compiling it into a lookup table for BSA, as summarized in Alg.~\ref{alg:table} and shown in Fig.~\ref{fig:overview}~(c). We estimate $P_h$ from a few calibration prompts by averaging attention over query--key pairs that share the same 3D offset, at a few evenly spaced \textit{step slots}, and each step reuses its nearest slot.
BSA operates on fixed-size key tiles, yet heads prefer different geometries, \eg, temporal columns or in-frame patches, so a box shape shared across heads as in VSA~\cite{vsa} covers little prior mass.
We instead make the tiling \textbf{\emph{head-adaptive}}: each head picks offline the 3D box shape that retains the most prior mass and packs its tokens into key tiles accordingly.
The table then stores $\mathrm{order}_h(a)$, the key tiles of each query tile $a$ ranked by prior mass, so inference takes the top-ranked positional tiles under budget $\rho$ by table lookup, without online scoring. App.~\ref{app:impl} gives the implementation details of this compilation.

\paragraph{Constructing the Content Residual.}
As discussed in Sec.~\ref{sec:motivation}, the content-dependent part of the residual is concentrated on a few input-specific keys and cannot be tabulated, so we locate it online. To keep this online selection efficient, the queries of a tile share one score: for query tile $a$ with pooled query $\bar{\mathbf q}_a$, key $j$ receives the logit $\bar{\mathbf q}_a^{\top}\mathbf k_j$, and the highest-scoring keys outside the positional tiles are selected as content keys and gathered into tiles.

\begin{figure}[t]
  \centering
  \includegraphics[width=\columnwidth]{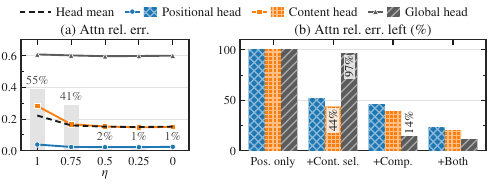}
  \vspace{-2em}
    \caption{\textbf{Selection and compensation are complementary.}
    (a) Single-layer attention error without compensation, versus ratio $\eta$; grey bars show the share of heads assigned each $\eta$.
    (b) Error left relative to the positional table alone; compensation removes most of the error that selection leaves on global heads.}
  \label{fig:head_types}
  \vspace{-1.3em}
\end{figure}

\paragraph{Head-Wise Ratio and Joint Key Selection.}
Following Insight~2, we determine the per-head positional ratio $\eta_h$ offline
on calibration prompts and reuse it for unseen inputs. Allocating a larger ratio
to online content selection reduces the relative attention output error, while
enabling online selection introduces additional scoring overhead. We restrict
$\eta_h$ to a discrete ladder $\{1, 0.75, 0.5, 0.25, 0\}$, representing the
ratio of selected keys drawn from the positional table.
Fig.~\ref{fig:head_types}~(a) shows diminishing error reduction as the online
ratio increases, with most gains achieved at $\eta_h=0.75$. We therefore target
80\% of the maximum attainable mean error reduction, starting from $\eta_h=1$
and greedily lowering one head by the ladder level to maximize the marginal
error reduction. App.~\ref{app:impl} gives the detailed algorithm. At inference, each query tile then takes an $\eta_h$ fraction
of its budget from its top-ranked positional tiles and fills the rest with the
highest-scoring content keys.

\paragraph{Discarded-Key Compensation.}
The selection above still discards every key outside the budget, and the error this leaves differs across heads.
Based on how the error of each head responds to the ratio $\eta_h$ in the allocation above, as shown in Fig.~\ref{fig:head_types}~(a), we characterize three head types: \emph{positional heads} are accurate with the positional prior table alone, \emph{content heads} start with a high error that content selection reduces, and \emph{global heads} start high and do not respond to selection. As shown in Fig.~\ref{fig:head_types}~(b), for global heads our best selection still leaves 97\% of the error. This is the thin tail over the discarded keys discussed in Sec.~\ref{sec:motivation}, which no selection within the budget can recover, so we approximate its aggregate contribution instead.
We condense the discarded keys of each query tile $a$ into a few \emph{virtual keys}, using the same pooled-query logits as content selection, $w_j=\exp(\bar{\mathbf q}_a^\top \mathbf k_j/\sqrt{d})$.
The discarded keys are split into cells, and each cell $c$ is pooled into
\begin{equation}
\tilde{\mathbf k}_c=\frac{\sum_{j\in c}w_j\,\mathbf k_j}{\sum_{j\in c}w_j},\qquad
\tilde{\mathbf v}_c=\frac{\sum_{j\in c}w_j\,\mathbf v_j}{\sum_{j\in c}w_j}.
\label{eq:comp}
\end{equation}
Each query attends to the selected keys and the virtual keys in one online softmax, where virtual key $c$ has logit $\mathbf q_i^{\top}\tilde{\mathbf k}_c/\sqrt{d}+b_c$ and the bias $b_c=\log\sum_{j\in c}w_j-\bar{\mathbf q}_a^{\top}\tilde{\mathbf k}_c/\sqrt{d}$ gives the cell's total attention mass.
For global heads, compensation removes 86\% of the error that selection leaves.

All three parts, positional tiles, content tiles, and compensation, are
implemented in one customized kernel with a single online-softmax pass,
detailed in App.~\ref{app:impl}.

\begin{figure}[t]
\centering
\includegraphics[width=\columnwidth]{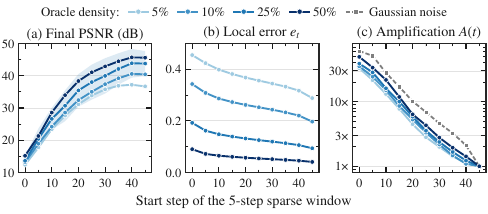}
\vspace{-2em}
\caption{\textbf{Step-dependent error and amplification.} One 5-step window runs oracle block-sparse attention; all other steps are dense. (a) Final PSNR. Higher is better. (b) Local attention error induced in the window. (c) Amplification of that error to the final output; Gaussian noise follows the same decay.}
\label{fig:step_tolerance}
\vspace{-1em}
\end{figure}

\subsection{Step-Adaptive Budget Allocation}
\label{sec:step}

\paragraph{Step-Dependent Error Amplification.}
To investigate the step-dependent sensitivity to sparsification, we apply magnitude-based oracle BSA to a single 5-step denoising window while keeping all remaining steps dense. As shown in Fig.~\ref{fig:step_tolerance}~(a), sparsifying earlier steps leads to greater degradation in final video quality. This behavior stems from two factors: sparse attention incurs larger relative error locally (Fig.~\ref{fig:step_tolerance}~(b)), and perturbations injected at earlier steps undergo stronger amplification during subsequent denoising (Fig.~\ref{fig:step_tolerance}~(c)). Gaussian noise injection exhibits similar amplification, indicating that the amplification is a property of the model and sampler rather than of sparse attention.

\begin{insight}{Insight 3}
Early-step attention errors are larger and far more amplified along the denoising trajectory, so the budget should be front-loaded toward early steps.
\end{insight}

\paragraph{Modeling Error Propagation.}
We model the final error as the per-step attention error weighted by its amplification, $\mathcal{E}=\sum_{t} A(t)\,e_t(\rho_t)$, where $e_t(\rho)$ is the local attention error at step $t$ and density $\rho$ on the calibration prompts. We measure $A(t)=D_T/D_t$ once per model, where $D_t$ is the latent deviation after step $t$ between a dense trajectory and one sparsified only at the first sparse step. After that step, the deviation grows by amplification alone. Across equal-cost budget schedules, the proxy $\mathcal{E}$ reliably predicts final video quality in Sec.~\ref{sec:ablation}, validating it as a surrogate objective.

\paragraph{Optimal Step-Wise Budget Allocation.}
With $\mathcal{E}$ as the surrogate objective, we formulate step-wise budget allocation as a constrained optimization problem instead of relying on hand-crafted schedules.
Specifically, given an average sparsity budget $\bar\rho$, we assign one density $\rho_s$ to each step slot of Sec.~\ref{sec:poco} and minimize $\mathcal{E}$ over the schedule:
\begin{equation}
\min_{\rho_s\in\mathcal{R}}\ \sum_{s} n_s A_s\, e_s(\rho_s)
\quad \text{s.t.}\quad \sum_{s} n_s\,\rho_s \le \bar\rho \sum_{s} n_s ,
\label{eq:budget}
\end{equation}
where slot $s$ serves $n_s$ steps, $A_s$ and $e_s$ are the amplification and local error averaged over them, and $\mathcal{R}$ is a discrete density ladder capped at $\rho^\ast$, where a sparse call becomes as slow as a dense one (App.~\ref{app:results}).
We solve Eq.~\eqref{eq:budget} exactly by dynamic programming, as detailed in Alg.~\ref{alg:budget} of App.~\ref{app:impl}.

%% file: sections/05_experiments_v2.tex
\section{Experiments}

\subsection{Experimental Setup}
\label{sec:setup}
\paragraph{Models and Tasks.}
We evaluate (1) text-to-video (T2V) tasks on Wan2.1-14B~\cite{wan}, HunyuanVideo-13B~\cite{hunyuanvideo}, and LTX-2.3-22B~\cite{ltx2} in its \textit{distilled} two-stage setting~\cite{sol}, (2) image-to-video (I2V) tasks on Wan2.1, and (3) video-to-video (V2V) editing tasks using Bernini~\cite{bernini}.
Following SVG-EAR~\cite{svgear}, we evaluate on 50 Penguin Benchmark prompts~\cite{hunyuanvideo} extended by the official Wan2.1 prompt extender~\cite{wan} for T2V, 50 VBench image--prompt pairs~\cite{vbenchpp} for I2V, and 73 EditVerseBench cases~\cite{editverse} for V2V.
PoCoAttn is calibrated once per model and task on 8 MovieGenBench prompts~\cite{moviegen}, VBench image--prompt pairs, or OpenVE-Bench videos~\cite{openve}, all disjoint from the evaluation sets.

\input{tables/main_table}

\paragraph{Baselines.}
We compare with FlashAttention-3 (FA3)~\cite{fa3} and, as applicable to each model, the static RadialAttn~\cite{radial}, the static and offline-calibrated CalibAtt~\cite{calibatt}, the dynamic SVG2~\cite{svg2}, SVG-EAR~\cite{svgear}, PISA~\cite{pisa}, and Sol-Attn~\cite{sol}. 
Each baseline uses its official implementation at a density no lower than ours and the same dense warm-up. 
Dense attention and all baselines except SVG2 and SVG-EAR, which keep their official backend with their own fused operators, share the backend with our fused non-attention operators.
All experiments run on NVIDIA H100 GPUs except the B200 results in Sec.~\ref{sec:efficiency}. App.~\ref{app:setup} gives further details.

\paragraph{Metrics.}
Following Sol-Attn~\cite{sol}, we report eight VBench~\cite{vbench} dimensions and their mean (AVG) for perceptual quality, and PSNR, SSIM, and LPIPS~\cite{lpips} for fidelity to dense-attention outputs.
For efficiency, we report the attention density, the end-to-end latency including condition encoding and VAE decoding (E2E), the DiT denoising latency including warm-up (DiT), and the attention and DiT speedups over FA3 (Attn$\times$ and Speedup).

\subsection{Main Results}
\paragraph{Video Generation.}
Tab.~\ref{tab:main} compares quality, fidelity, and efficiency across the evaluated video models.
On Wan2.1-T2V, PoCoAttn at 25\% density improves PSNR by 3.4~dB over the strongest baseline, PISA at 30\% density, and runs faster, with a 1.40$\times$ speedup against 1.19$\times$.
The fully offline CalibAtt~\cite{calibatt} reaches the highest attention speedup among the baselines, 2.27$\times$ at 36.6\% density, while PoCoAttn at 15\% density is both 2.2~dB more faithful and faster, with a 3.09$\times$ attention speedup. Pushing the density to 10\%, PoCoAttn reaches a 4.14$\times$ attention
speedup and a 1.73$\times$ DiT speedup at 25.07\,dB PSNR,
still above every baseline in Tab.~\ref{tab:main}.
On Wan2.1-I2V, HunyuanVideo, and LTX-2.3, PoCoAttn likewise achieves the best fidelity among the sparse methods, improving PSNR by 1.9--5.0~dB over the strongest baseline with 1.39--1.65$\times$ speedup at 25\% density and 1.60--2.02$\times$ at 15\%, while it outperforms or remains comparable to dense attention and the other baselines on VBench.
As Fig.~\ref{fig:teaser}~(b) shows on Wan2.1-T2V, the baselines trade fidelity for speed along a shared frontier, while PoCoAttn is \textit{\textbf{Pareto-optimal}.}
The gains hold across prompts: at 25\% density, PoCoAttn outperforms the strongest baseline on at least 48 of the 50 prompts in every setting. More results are provided in App.~\ref{app:results}.

\begin{figure*}[t]
  \centering
  \includegraphics[width=\textwidth]{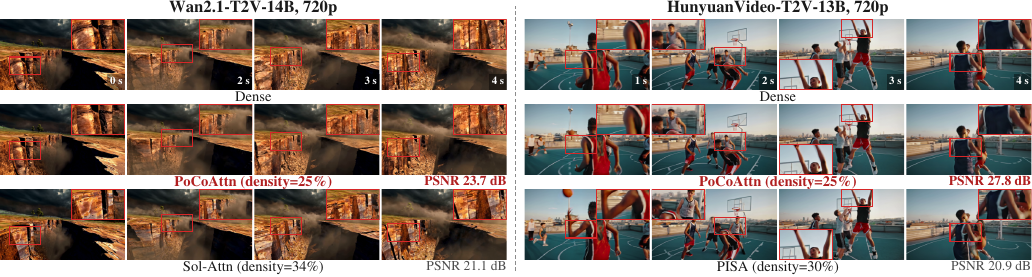}
  \vspace{-2em}
  \caption{\textbf{Qualitative comparison of video generation.} Our method achieves better visual fidelity at lower density.}
  \label{fig:qualitative}
  \vspace{-1em}
\end{figure*}

\paragraph{Qualitative Comparison.} 
Fig.~\ref{fig:qualitative} shows the qualitative comparison of four video frames between baselines and PoCoAttn at lower density. As highlighted in the red-boxed close-up, Sol-Attn loses the rock texture of the canyon walls, and PISA alters the positions and poses of the basketball players, whereas PoCoAttn preserves the details, motion, and the global layout of the dense output. Additional qualitative results are provided in App.~\ref{app:qual}.

\paragraph{Video Editing.}
Beyond T2V and I2V, we evaluate PoCoAttn on video editing with Bernini~\cite{bernini}, where attention spans both the source and the edited video. As shown in Tab.~\ref{tab:bernini}, PoCoAttn at 25\% density improves PSNR by 3.9~dB over the strongest baseline, at a higher speedup, and wins on at least 70 of the 73 cases against either baseline; at 15\% density, it remains 2.2~dB higher with a 1.72$\times$ speedup.
\input{tables/Bernini}

\paragraph{Results on Blackwell GPUs.}
To assess generalization across GPU architectures, we evaluate PoCoAttn on the NVIDIA B200, porting our Triton kernel to CuTe DSL and using FlashAttention-4~\cite{fa4} as the dense reference. As shown in Tab.~\ref{tab:b200}, PoCoAttn achieves 27.48~dB PSNR at 25\% density with a 1.44$\times$ speedup, whereas Sol-Attn and PISA reach only 0.99$\times$ and 1.03$\times$: the official Blackwell kernel of Sol-Attn does not yet fully exploit the hardware, as its authors note~\cite{sol}, and PISA has no official Blackwell implementation. The gains extend across models, reaching 1.56$\times$ on LTX-2.3 and 1.61$\times$ on HunyuanVideo.

\input{tables/Blackwell}

\paragraph{Efficiency Analysis.}
\label{sec:efficiency}
Fusing the non-attention operators alone brings a 1.04--1.26$\times$ E2E speedup over the original dense pipeline, and adding PoCoAttn at 15\% density raises it to 1.85--1.99$\times$, with the resulting latency listed in Tab.~\ref{tab:main}. Calibration is a one-time offline cost: 44 H100 GPU-hours on 8 prompts for Wan2.1-14B at 720p, and the resulting calibration supports all tested density settings. Moreover, even a single calibration prompt suffices, which further reduces the calibration cost: on LTX-2.3, one prompt matches the quality of 8 and takes only about 7 H100 GPU-hours.
The tables also \textbf{\textit{transfer without recalibration}}, from 480p to 720p with a 0.28\,dB loss and from T2V to I2V on par with the native ones (Tab.~\ref{tab:calib_transfer}). App.~\ref{app:results} details the latency breakdown, kernel roofline, calibration cost and memory overhead.

\subsection{Ablation Studies}
\label{sec:ablation}

\paragraph{Component Ablation.}

Tab.~\ref{tab:abl_component} ablates the three components of PoCoAttn at 25\% density on Wan2.1-T2V-14B at 720p, with all rows using the same solved step-wise budget.
Positional keys alone reach 24.12~dB with no online selection cost and the highest attention speedup of 2.38$\times$.
Content keys alone are 1.2~dB better but barely improve SSIM, whereas combining the two, with only about 13\% of the keys selected online, yields 26.11~dB and 0.850 SSIM: the positional prior supplies the stable structure, and online selection is best spent on the few keys it misses.
Compensation adds 1.7~dB and reduces LPIPS from 0.112 to 0.081 without selecting any additional key. The remaining error thus lies in the diffuse mass of the discarded keys, consistent with the global heads in Fig.~\ref{fig:head_types}~(b), and removing it costs about 9\% of the attention speedup, or only 0.03$\times$ in denoising speedup.
Together the components address the input-stable positional prior, the input-dependent content residual, and the diffuse tail over the discarded keys, raising PSNR by 3.7~dB over position-only and 2.5~dB over content-only selection while retaining a 1.40$\times$ denoising speedup.

\input{tables/ablation_poco}

\input{tables/ablation_budget_alloc}

\paragraph{Step-Schedule Ablation.}

Tab.~\ref{tab:sched_ablation} compares four strategies for allocating a fixed 25\% average sparsity budget across denoising steps on T2V, 720p. All schedules run at the same speed within noise, since the latency of the sparse kernel is linear in density, as shown in App.~\ref{app:results}. The error proxy $\mathcal{E}$, estimated offline as described in Sec.~\ref{sec:step}, predicts the ranking of the measured fidelity, validating it as an optimization objective. The reversed schedule, which follows the observation from token pruning~\cite{diffratio} that early steps tolerate more compression, falls 2.7~dB below even the uniform schedule, confirming that sparse attention exhibits the opposite trend. Our solved schedule improves PSNR by 0.35~dB over the hand-crafted schedule and by 1.9~dB over the uniform one, which itself already outperforms the strongest baseline in Tab.~\ref{tab:main} by 1.5~dB, so both the position--content selection and the step-wise allocation contribute substantially.

%% file: tables/main_table.tex
\providecolor{oursred}{RGB}{255,236,236}
\begin{table*}[t]
\centering
\vspace{-1em}
\caption{\textbf{Quantitative comparison of video generation quality and efficiency.}
Bold indicates the best sparse-method results. Density is the fraction of query--key pairs actually computed in the sparse attention calls; percentages in parentheses are target density settings. $^\dagger$Lower speedup than originally reported, as our fused dense baseline is faster than their official dense run (App.~\ref{app:setup}).}
\label{tab:main}
\vspace{-0.5em}
\small
\setlength{\tabcolsep}{3pt}
\renewcommand{\arraystretch}{0.9}
\setlength{\aboverulesep}{0.15ex}
\setlength{\belowrulesep}{0.25ex}
\resizebox{\textwidth}{!}{%
\begin{NiceTabular}{l|ccccccccc|ccc|ccccc}
\CodeBefore
  \rowcolor{oursred}{11,12,19,20,26,27,34,35}
\Body
\toprule
\multirow{2}{*}{\textbf{Method}}
& \multicolumn{9}{c|}{\textbf{\boldmath VBench$\uparrow$}} & \multicolumn{3}{c|}{\textbf{Fidelity}} & \multicolumn{5}{c}{\textbf{Efficiency}} \\
\cmidrule(lr){2-10}\cmidrule(lr){11-13}\cmidrule(lr){14-18}
& \textbf{\boldmath SC}
& \textbf{\boldmath BC}
& \textbf{\boldmath TF}
& \textbf{\boldmath MS}
& \textbf{\boldmath AQ}
& \textbf{\boldmath IQ}
& \textbf{\boldmath DD}
& \textbf{\boldmath OC}
& \textbf{\boldmath AVG}
& \textbf{\boldmath PSNR$\uparrow$}
& \textbf{\boldmath SSIM$\uparrow$}
& \textbf{\boldmath LPIPS$\downarrow$}
& \textbf{\boldmath Density$\downarrow$}
& \textbf{\boldmath E2E (s)$\downarrow$}
& \textbf{\boldmath DiT (s)$\downarrow$}
& \textbf{\boldmath Attn$\times\uparrow$}
& \textbf{\boldmath Speedup$\uparrow$} \\
\midrule
\multicolumn{18}{c}{\textbf{Wan2.1-T2V-14B} \textit{(14B, Text-to-Video, 720p, 81 Frames)}} \\
\midrule
Dense (FA3) & 92.32 & 95.35 & 97.42 & 98.54 & 61.17 & 64.14 & 56.00 & 24.47 & 73.68 & -- & -- & -- & 100\% & 963.2 & 954.9 & 1.00$\times$ & 1.00$\times$ \\
RadialAttn~\cite{radial} & 92.32 & 95.35 & 97.33 & 98.48 & 61.70 & 63.90 & 56.00 & 24.47 & 73.69 & 22.50 & 0.775 & 0.188 & 26.4\% & 743.1 & 735.0 & 1.67$\times$ & 1.30$\times$ \\
SVG2~\cite{svg2}$^\dagger$ & 92.04 & 95.28 & 97.49 & 98.54 & 61.27 & 63.31 & 52.00 & 24.38 & 73.04 & 23.60 & 0.803 & 0.155 & 31.2\% & 869.9 & 861.6 & 1.40$\times$ & 1.11$\times$ \\
SVG-EAR~\cite{svgear}$^\dagger$ & 92.00 & 95.21 & 97.47 & 98.55 & 61.34 & 63.66 & 54.00 & 24.34 & 73.32 & 24.28 & 0.819 & 0.140 & 27.0\% & 883.1 & 874.4 & 1.33$\times$ & 1.09$\times$ \\
PISA~\cite{pisa} & 92.22 & 95.11 & 97.38 & 98.53 & 61.15 & 64.21 & 56.00 & 24.31 & 73.61 & 24.42 & 0.828 & 0.134 & 30.0\% & 808.4 & 800.3 & 1.41$\times$ & 1.19$\times$ \\
Sol-Attn~\cite{sol} & 92.48 & 95.36 & 97.40 & 98.51 & 61.54 & 64.13 & 54.00 & 24.46 & 73.49 & 24.12 & 0.822 & 0.140 & 33.8\% & 739.2 & 730.9 & 1.74$\times$ & 1.31$\times$ \\
CalibAtt~\cite{calibatt} & 91.96 & 95.22 & 97.30 & 98.50 & 60.95 & 64.23 & 50.00 & 24.35 & 72.81 & 24.20 & 0.816 & 0.139 & 36.6\% & 676.3 & 667.5 & 2.27$\times$ & 1.43$\times$ \\
\textbf{PoCoAttn} (25\%) & 92.36 & 95.50 & 97.42 & 98.52 & 61.41 & 64.15 & 58.00 & 24.36 & \textbf{73.97} & \textbf{27.81} & \textbf{0.884} & \textbf{0.081} & 25.0\% & 690.9 & 682.7 & 2.07$\times$ & 1.40$\times$ \\
\textbf{PoCoAttn} (15\%) & 92.34 & 95.44 & 97.44 & 98.53 & 61.19 & 63.83 & 58.00 & 24.49 & 73.91 & 26.36 & 0.858 & 0.105 & 15.0\% & \textbf{607.2} & \textbf{598.9} & \textbf{3.09$\times$} & \textbf{1.59$\times$} \\
\midrule
\multicolumn{18}{c}{\textbf{Wan2.1-I2V-14B} \textit{(14B, Image-to-Video, 720p, 81 Frames)}} \\
\midrule
Dense (FA3) & 95.36 & 96.13 & 96.56 & 98.10 & 63.50 & 70.89 & 48.00 & 26.77 & 74.41 & -- & -- & -- & 100\% & 987.5 & 973.7 & 1.00$\times$ & 1.00$\times$ \\
SVG2~\cite{svg2}$^\dagger$ & 95.25 & 95.84 & 96.56 & 98.05 & 63.41 & 70.77 & 46.00 & 26.81 & 74.09 & 24.40 & 0.796 & 0.114 & 30.5\% & 874.2 & 860.6 & 1.45$\times$ & 1.13$\times$ \\
SVG-EAR~\cite{svgear}$^\dagger$ & 95.36 & 95.96 & 96.58 & 98.08 & 63.58 & 70.82 & 48.00 & 26.81 & 74.40 & 25.52 & 0.816 & 0.099 & 26.6\% & 890.4 & 877.5 & 1.36$\times$ & 1.11$\times$ \\
PISA~\cite{pisa} & 95.14 & 95.79 & 96.51 & 98.09 & 63.47 & 70.77 & 48.00 & 26.85 & 74.33 & 26.37 & 0.858 & 0.079 & 30.0\% & 833.9 & 820.8 & 1.41$\times$ & 1.19$\times$ \\
Sol-Attn~\cite{sol} & 95.17 & 95.94 & 96.51 & 98.09 & 63.59 & 70.76 & 48.00 & 26.82 & 74.36 & 25.12 & 0.819 & 0.104 & 33.5\% & 750.7 & 737.7 & 1.78$\times$ & 1.32$\times$ \\
\textbf{PoCoAttn} (25\%) & 95.26 & 96.09 & 96.58 & 98.11 & 63.51 & 70.98 & 48.00 & 26.73 & 74.41 & \textbf{28.38} & \textbf{0.883} & \textbf{0.067} & 25.0\% & 711.8 & 698.3 & 2.09$\times$ & 1.39$\times$ \\
\textbf{PoCoAttn} (15\%) & 95.20 & 96.02 & 96.62 & 98.12 & 63.62 & 71.07 & 48.00 & 26.79 & \textbf{74.43} & 27.24 & 0.862 & 0.083 & 15.0\% & \textbf{620.3} & \textbf{607.1} & \textbf{3.17$\times$} & \textbf{1.60$\times$} \\
\midrule
\multicolumn{18}{c}{\textbf{HunyuanVideo} \textit{(13B, Text-to-Video, 720p, 129 Frames)}} \\
\midrule
Dense (FA3) & 91.58 & 95.51 & 98.35 & 99.18 & 60.66 & 61.12 & 54.00 & 23.95 & 73.04 & -- & -- & -- & 100\% & 953.2 & 924.6 & 1.00$\times$ & 1.00$\times$ \\
SVG2~\cite{svg2}$^\dagger$ & 91.45 & 95.45 & 98.38 & 99.14 & 59.75 & 60.64 & 48.00 & 24.01 & 72.10 & 29.69 & 0.903 & 0.078 & 26.6\% & 769.6 & 741.1 & 1.67$\times$ & 1.25$\times$ \\
PISA~\cite{pisa} & 91.59 & 95.41 & 98.30 & 99.18 & 60.30 & 61.51 & 52.00 & 23.97 & 72.78 & 30.17 & 0.912 & 0.072 & 30.2\% & 781.9 & 753.3 & 1.37$\times$ & 1.23$\times$ \\
Sol-Attn~\cite{sol} & 91.49 & 95.61 & 98.31 & 99.19 & 60.73 & 61.38 & 56.00 & 23.99 & \textbf{73.34} & 28.83 & 0.898 & 0.085 & 32.6\% & 671.1 & 642.3 & 1.86$\times$ & 1.44$\times$ \\
\textbf{PoCoAttn} (25\%) & 91.49 & 95.58 & 98.36 & 99.18 & 60.46 & 60.97 & 52.00 & 24.08 & 72.77 & \textbf{35.17} & \textbf{0.954} & \textbf{0.033} & 25.1\% & 641.4 & 612.6 & 2.01$\times$ & 1.51$\times$ \\
\textbf{PoCoAttn} (15\%) & 91.49 & 95.63 & 98.37 & 99.18 & 60.62 & 60.92 & 52.00 & 24.05 & 72.78 & 32.57 & 0.934 & 0.052 & 15.0\% & \textbf{544.3} & \textbf{515.9} & \textbf{2.93$\times$} & \textbf{1.79$\times$} \\
\midrule
\multicolumn{18}{c}{\textbf{LTX-2.3} \textit{(22B, distilled, Text-to-Video, 1080p, 721 Frames)}} \\
\midrule
Dense (FA3) & 83.57 & 92.10 & 98.88 & 99.43 & 52.45 & 48.33 & 60.00 & 22.13 & 69.61 & -- & -- & -- & 100\% & 193.2 & 168.7 & 1.00$\times$ & 1.00$\times$ \\
RadialAttn~\cite{radial} & 75.50 & 89.28 & 98.92 & 99.34 & 42.61 & 39.90 & 58.00 & 18.20 & 65.22 & 24.54 & 0.765 & 0.346 & 26.2\% & 139.4 & 115.0 & 1.82$\times$ & 1.47$\times$ \\
PISA~\cite{pisa} & 83.54 & 91.91 & 98.87 & 99.42 & 52.68 & 49.27 & 60.00 & 22.13 & 69.73 & 31.14 & 0.880 & 0.123 & 30.0\% & 167.8 & 143.3 & 1.27$\times$ & 1.18$\times$ \\
Sol-Attn~\cite{sol} (30\%) & 83.49 & 92.07 & 98.88 & 99.42 & 52.76 & 48.86 & 60.00 & 22.04 & 69.69 & 32.10 & 0.892 & 0.108 & 31.3\% & 158.7 & 134.1 & 1.41$\times$ & 1.26$\times$ \\
Sol-Attn~\cite{sol} (15\%) & 83.68 & 92.07 & 98.89 & 99.43 & 53.52 & 49.82 & 60.00 & 22.21 & 69.95 & 29.66 & 0.861 & 0.152 & 15.4\% & 114.6 & 90.1 & 2.99$\times$ & 1.87$\times$ \\
\textbf{PoCoAttn} (25\%) & 83.74 & 92.04 & 98.87 & 99.44 & 53.37 & 49.86 & 60.00 & 22.33 & 69.96 & \textbf{33.95} & \textbf{0.917} & \textbf{0.081} & 25.0\% & 126.8 & 102.4 & 2.27$\times$ & 1.65$\times$ \\
\textbf{PoCoAttn} (15\%) & 83.81 & 92.20 & 98.86 & 99.44 & 53.77 & 50.66 & 60.00 & 22.47 & \textbf{70.15} & 31.69 & 0.893 & 0.111 & 15.0\% & \textbf{108.0} & \textbf{83.4} & \textbf{3.48$\times$} & \textbf{2.02$\times$} \\
\bottomrule
\CodeAfter
\tikz \draw[line width=\arrayrulewidth]
    (1-|2) -- (3-|2)
    (1-|11) -- (3-|11)
    (1-|14) -- (3-|14);
\end{NiceTabular}%
}
\vspace{-1em}
\end{table*}

%% file: tables/Bernini.tex
\begin{table}[t]
\centering
\caption{\textbf{Video editing on Bernini.} Speedup over dense FA3.}
\label{tab:bernini}
\vspace{-3mm}
\footnotesize
\setlength{\tabcolsep}{3pt}
\renewcommand{\arraystretch}{0.9}
\begin{tabular}{l|ccccc}
\toprule
\textbf{Method}
& \textbf{PSNR}$\uparrow$
& \textbf{SSIM}$\uparrow$
& \textbf{LPIPS}$\downarrow$
& \textbf{Density}$\downarrow$
& \textbf{Speedup}$\uparrow$ \\
\midrule
PISA~\cite{pisa} & 33.72 & 0.949 & 0.032 & 25.0\% & 1.28$\times$ \\
Sol-Attn~\cite{sol} & 33.35 & 0.947 & 0.033 & 26.9\% & 1.47$\times$ \\
\rowcolor{oursred}
\textbf{PoCoAttn} (25\%) & \textbf{37.60} & \textbf{0.968} & \textbf{0.017} & 25.0\% & 1.47$\times$ \\
\rowcolor{oursred}
\textbf{PoCoAttn} (15\%) & 35.90 & 0.961 & 0.024 & 15.0\% & \textbf{1.72$\times$} \\
\bottomrule
\end{tabular}
\vspace{-1.3em}
\end{table}

%% file: tables/Blackwell.tex
\begin{table}[t]
\centering

\caption{\textbf{Results on the NVIDIA B200} (Wan2.1-T2V-14B, 720p).}
\vspace{-3mm}
\label{tab:b200}
\footnotesize
\setlength{\tabcolsep}{3pt}
\renewcommand{\arraystretch}{0.92}
\begin{tabular}{l|ccccc}
\toprule
\textbf{Method}
& \textbf{PSNR}$\uparrow$
& \textbf{SSIM}$\uparrow$
& \textbf{LPIPS}$\downarrow$
& \textbf{Density}$\downarrow$
& \textbf{Speedup}$\uparrow$ \\
\midrule
Dense (\textbf{FA4}~\cite{fa4}) & -- & -- & -- & 100\% & 1.00$\times$ \\
PISA~\cite{pisa} & 23.71 & 0.805 & 0.154 & 29.9\% & 1.03$\times$ \\
Sol-Attn~\cite{sol} & 23.85 & 0.806 & 0.152 & 30.7\% & 0.99$\times$ \\
\rowcolor{oursred}
\textbf{PoCoAttn} & \textbf{27.48} & \textbf{0.876} & \textbf{0.088} & 25.0\% & \textbf{1.44$\times$} \\
\bottomrule
\end{tabular}
\vspace{-1.5em}
\end{table}

%% file: tables/ablation_poco.tex
\providecommand{\cmark}{\checkmark}

\begin{table}[t]
\centering

\caption{\textbf{Component ablation} on Wan2.1. Pos.: positional keys from the table; Cont.: online content keys; Comp.: compensation.}
\vspace{-0.8em}
\label{tab:abl_component}
\footnotesize
\setlength{\tabcolsep}{2pt}
\resizebox{\columnwidth}{!}{%
\begin{tabular}{ccc|ccccc}
\toprule
\textbf{Pos.} & \textbf{Cont.} & \textbf{Comp.}
& \textbf{PSNR}$\uparrow$
& \textbf{SSIM}$\uparrow$
& \textbf{LPIPS}$\downarrow$
& \textbf{Attn}$\times$ $\uparrow$
& \textbf{Speedup}$\uparrow$ \\
\midrule
\cmark &        &        & 24.12 & 0.811 & 0.149 & \textbf{2.38$\times$} & \textbf{1.46$\times$} \\[-1pt]
       & \cmark &        & 25.30 & 0.818 & 0.135 & 2.24$\times$ & 1.42$\times$ \\[-1pt]
\cmark & \cmark &        & 26.11 & 0.850 & 0.112 & 2.27$\times$ & 1.43$\times$ \\[-1pt]
\rowcolor{oursred}
\cmark & \cmark & \cmark & \textbf{27.81} & \textbf{0.884} & \textbf{0.081} & 2.07$\times$ & 1.40$\times$ \\[-1pt]
\bottomrule
\end{tabular}%
}
\vspace{-0.7em}
\end{table}

%% file: tables/ablation_budget_alloc.tex
\newcommand{\ladderw}{0.45em}   %
\newcommand{\ladderh}{2.4ex}    %
\newcommand{\ladder}[1]{%
  \begin{tikzpicture}[baseline=-0.2ex,x=\ladderw,y=\ladderh]
    \draw[gray!60,line width=0.3pt] (0,0) -- (8,0);
    \foreach \d [count=\i from 0] in {#1}{
      \fill[black!75] (\i+0.08,0) rectangle (\i+0.92,\d/0.6);
    }
  \end{tikzpicture}%
}

\begin{table}[t]
\centering

\caption{\textbf{Step-schedule ablation} on Wan2.1. All rows run PoCoAttn at 25\% average density and differ only in how the budget is allocated across denoising steps (bars: early$\rightarrow$late).
$\mathcal{E}$ is the error proxy in Sec.~\ref{sec:step}; Solved is our schedule that minimizes it.}
\vspace{-2mm}
\label{tab:sched_ablation}
\footnotesize
\setlength{\tabcolsep}{2pt}
\resizebox{\columnwidth}{!}{%
\begin{tabular}{lcc|cccc}
\toprule
\textbf{Schedule}
& \textbf{Density}
& \textbf{\boldmath$\mathcal{E}$}$\downarrow$
& \textbf{PSNR}$\uparrow$
& \textbf{SSIM}$\uparrow$
& \textbf{LPIPS}$\downarrow$
& \textbf{Speedup}$\uparrow$ \\
\midrule
Uniform      & \ladder{.25,.25,.25,.25,.25,.25,.25,.25} & 2.18 & 25.94 & 0.857 & 0.105 & 1.42$\times$ \\
Reversed     & \ladder{.05,.05,.10,.15,.20,.40,.45,.45} & 4.63 & 23.23 & 0.795 & 0.168 & 1.41$\times$ \\
Hand-crafted & \ladder{.45,.45,.45,.20,.20,.20,.10,.10} & 1.50 & 27.46 & 0.881 & 0.083 & 1.40$\times$ \\
\rowcolor{oursred}
\textbf{Solved (ours)} & \ladder{.55,.55,.50,.25,.15,.10,.05,.05} & \textbf{1.34} & \textbf{27.81} & \textbf{0.884} & \textbf{0.081} & 1.40$\times$ \\
\bottomrule
\end{tabular}%
}
\vspace{-1.5em}
\end{table}

%% file: sections/06_conclusion.tex
\vspace{-0.5em}
\section{Conclusion}
We presented PoCoAttn, a training-free sparse attention framework that resolves the input-stable positional prior of video attention offline and its content residual online, and allocates the budget across denoising steps by error amplification.
Across four video diffusion models and three tasks, it consistently advances the speed--fidelity Pareto frontier.
PoCoAttn offers a direction for efficient video generation that combines static and dynamic sparse attention methods.

%% file: sections/07_appendix.tex
\clearpage
\maketitlesupplementary
\makeatletter\write\@auxout{\string\gdef\string\SuppStartPage{\the\ReadonlyShipoutCounter}}\makeatother
\appendix
\renewcommand{\thefigure}{S\arabic{figure}}\setcounter{figure}{0}
\renewcommand{\thetable}{S\arabic{table}}\setcounter{table}{0}

\section{Extended Related Work}\label{app:related}
\paragraph{Static Sparse Attention.}
Static methods derive their masks from a pattern family fixed before inference.
Template-based methods predefine patterns such as sliding 3D windows~\cite{sta}, distance-decaying bands~\cite{radial}, spatial and temporal patterns~\cite{svg}, or diagonal and stripe menus~\cite{sparsevdit}; MOD-DiT~\cite{moddit} further fits a linear mixture of such templates from the early denoising steps.
Motivated by empirical observations, these templates cover only part of the attention structure and may not generalize across models with distinct sparsity patterns. Some also select or fit the template at run time~\cite{svg,moddit}, yet the mask never leaves the predefined family.
Calibration-based methods instead exploit the stability of attention patterns across inputs, deriving masks offline and reusing them without online selection:
CalibAtt~\cite{calibatt} and ScalingAttn~\cite{scalingattn} aggregate input-stable block masks from full attention maps without examining the underlying structure, while PSA~\cite{psa} searches per-head parameters of a stripe family, which still fixes the pattern form.
Our analysis reveals the head-specific positional prior behind this stability, which generalizes both hand-crafted and parameterized patterns, whereas the content-dependent residual cannot be captured by any static mask.

\paragraph{Dynamic Sparse Attention.}
Dynamic methods select keys at run time by scoring blocks with cheap proxies~\cite{spargeattn,xattention,pisa,sol}, clustering similar tokens~\cite{svg2,svgear,svoo}, or ranking tokens by exact attention mass~\cite{mcsparse}. Trainable methods learn the selection end to end~\cite{vsa}. Recent work further compensates for the discarded entries to approximate the full attention output~\cite{pisa,sol,svgear,mcsparse}. Luo~\etal~\cite{svoo} additionally profiles layer-wise sparsity offline, but still constructs its content-adaptive blocks online. These methods adapt to content but rescore all keys at every call or anchor step, incurring substantial yet avoidable overhead to rediscover structure that is largely fixed by position. PoCoAttn draws most keys from the offline positional prior, searches online only for a small, head-specific share of content keys, and approximates the contribution of the unselected keys within the same online softmax rather than discarding it.

\section{Theory of Attention Decomposition}
\label{app:decomp}
As described in Sec.~\ref{prelim}, 3D RoPE encodes relative position into the attention logit $s_{ij}=q_i^{\top}R(\Delta_{ij})k_j/\sqrt{d}$ only through the relative offset $\Delta_{ij}$, so we look for an offline predictor of attention that depends on the offset alone. This section shows that the positional prior in Eq.~\eqref{eq:decomp} is the best such predictor for unseen inputs, and that offline calibration is a finite-sample estimate of it. We keep the notation of Eq.~\eqref{eq:decomp} for a fixed head $h$ and take all expectations uniformly over inputs and query--key pairs:
\begin{equation}
\begin{aligned}
P_h(\Delta)&=\mathbb{E}\big[A_{ij}\mid\Delta_{ij}=\Delta\big],\\
r_{ij}&=A_{ij}-P_h(\Delta_{ij}).
\end{aligned}
\label{eq:residual_def}
\end{equation}
An offset prior also has $O(N)$ entries for $N$ tokens rather than the $N^2$ of a per-pair table. The results below do not rely on RoPE and hold for any attention weights.

\begin{figure*}[t]
  \centering
  \includegraphics[width=\textwidth]{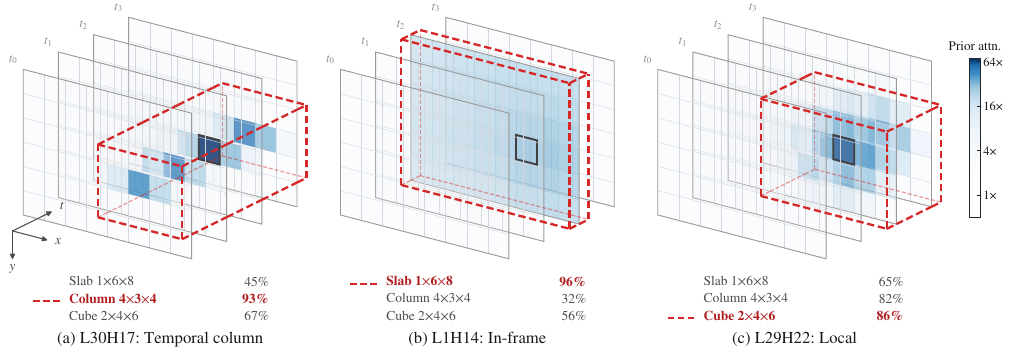}
  \caption{\textbf{Head-adaptive compilation.}
  Positional priors of three Wan2.1 heads for one query (black outline) on a toy $4\times6\times8$ grid.
  At a 25\% key budget, each head selects the box shape ($t\times y\times x$) among all candidates that retains the most prior mass (red dashed). Percentages give the retained mass of each candidate. The head-adaptive box shapes are decided offline and reused for table construction.}
  \label{fig:box_shapes}
\end{figure*}

\paragraph{Optimality.}
By definition of $P_h$, the residual has zero mean at every offset. It is therefore orthogonal to every function $g$ of the offset:
\begin{equation}
\mathbb{E}[r_{ij}\mid\Delta_{ij}]=0
\;\Longrightarrow\;
\mathbb{E}\big[r_{ij}\,g(\Delta_{ij})\big]=0.
\label{eq:orth}
\end{equation}
For any offset-only predictor $f$, taking $g=P_h-f$ gives
\begin{equation}
\begin{aligned}
\mathbb{E}\big[(A_{ij}-f(\Delta_{ij}))^2\big]
&=\underbrace{\mathbb{E}\big[r_{ij}^2\big]}_{\text{minimum error}}\\
&\quad+\underbrace{\mathbb{E}\big[(P_h(\Delta_{ij})-f(\Delta_{ij}))^2\big]}_{\text{excess error}}.
\end{aligned}
\label{eq:pythagoras}
\end{equation}
The excess error vanishes only when $f=P_h$, so $P_h$ is the optimal offset-only predictor of attention in mean squared error. The minimization decouples across offsets, so the unequal number of pairs per offset near the grid boundary does not affect the optimum. The same prior also gives the selection rule. An offset mask retains, in expectation, the sum of $P_h(\Delta)$ over its query--key pairs, so keeping the offsets with the largest $P_h$ retains the most attention mass at a given key budget. This is the rule behind the positional table in Sec.~\ref{sec:poco}, which ranks key tiles by the prior.

\begin{figure}[t]
  \centering
  \includegraphics[width=\columnwidth]{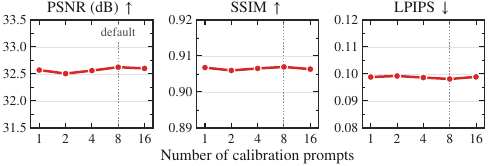}
  \caption{Effect of calibration set size on video fidelity, the 20-prompt subset evaluated on LTX-2.3-22B. Uniform 25\% density.}
  \label{fig:calib_size}
\end{figure}

\paragraph{Offline Calibration.}
The prior $P_h$ is a population quantity and is unknown. In practice it is estimated by averaging attention at each offset over a few calibration inputs. For brevity, the main text writes $P_h$ for both the prior and this estimate. In this section we distinguish the two and denote the estimate by $\hat P_h$. On the calibration inputs themselves, $\hat P_h$ is exactly optimal and its residual is exactly zero-mean at each offset. On unseen inputs, these properties hold only up to the calibration error below. For an input unseen during calibration, $\hat P_h$ is a fixed offset-only predictor, so Eq.~\eqref{eq:pythagoras} with $f=\hat P_h$ gives
\begin{equation}
\begin{aligned}
\mathbb{E}\big[(A_{ij}-\hat P_h(\Delta_{ij}))^2\big]
&=\underbrace{\mathbb{E}\big[r_{ij}^2\big]}_{\text{irreducible by offset}}\\
&\quad+\underbrace{\mathbb{E}\big[(P_h(\Delta_{ij})-\hat P_h(\Delta_{ij}))^2\big]}_{\text{calibration error}}.
\end{aligned}
\label{eq:calib_error}
\end{equation}
The first term cannot be reduced by any offline table and is left to online content key selection in Sec.~\ref{sec:poco}. The second term is the only cost of calibrating on finitely many inputs. It is small in practice, since priors estimated from individual prompts agree with a mean correlation of $0.92$ in Fig.~\ref{fig:attn_decomp}, and the retained mass estimated offline matches that measured on unseen prompts in Fig.~\ref{fig:prior_coverage}. Its effect on the generated videos is also negligible, as varying the number of calibration prompts from 1 to 16 changes PSNR by at most 0.12\,dB on LTX-2.3 in Fig.~\ref{fig:calib_size}.

\paragraph{Scope.}
The analysis above gives the theoretical basis of PoCoAttn: the positional prior is the optimal offset-only predictor of attention, and ranking by it retains the most attention mass among offset masks. Its stability across inputs is established empirically in Sec.~\ref{sec:motivation}, and together they allow the prior to be calibrated once offline. The part of attention that depends on the current input is handled by the content keys selected online. The prior is defined at the token level, and the positional table used at inference is compiled from it at the tile level, as detailed in App.~\ref{app:impl}.

\section{Implementation Details}\label{app:impl}

\paragraph{Calibration Runs.}
Each model is calibrated once under its evaluation setting.
Text-to-video models use eight MovieGenBench~\cite{moviegen} prompts disjoint from the evaluation sets, Wan2.1-I2V uses eight held-out VBench-I2V~\cite{vbenchpp} image--prompt pairs, and Bernini uses eight OpenVE-Bench~\cite{openve} videos.
A calibration run is a dense generation in which hooks capture the post-RoPE queries, keys, and values of the conditional attention calls without altering the trajectory.
Statistics are collected at a few step slots, and each sparse step uses the tables of its nearest slot.
Tab.~\ref{tab:deploy} lists the slots for the evaluated video generation models.

\paragraph{Head-Adaptive Compilation.}
As in Alg.~\ref{alg:table}, $P_h$ is measured on sampled query rows whose attention weights are computed exactly.
We set $G=64$, so a 128-row query tile covers two consecutive key tiles. The candidates $\mathcal{B}$ are all 3D boxes of at most $G$ tokens that fit the token grid. As the toy example in Fig.~\ref{fig:box_shapes} shows, each head selects the box shape that retains the most prior-predicted attention per query. On Wan2.1-T2V at 720p, the heads select 30 distinct shapes, led by the temporal column $21\times3\times1$ and the spatial patch $1\times4\times16$.
$\pi_h$ concatenates the boxes, with tokens in raster order within each box, and cuts the sequence every $G$ tokens, so a key tile may span two boxes and only the last tile is partial.
At inference, $\mathbf Q$, $\mathbf K$, and $\mathbf V$ are gathered into this order before attention so that the token layout is continuous, and the output is scattered back afterwards.

\input{tables/Settings.tex}

\paragraph{Step-Adaptive Budget Allocation.}
As shown in Fig.~\ref{fig:overview}, after table construction, two GPU passes provide all inputs.
The first runs the trajectories that measure the amplification profile $A(t)$ of Sec.~\ref{sec:step}.
The second is a probe that measures the single-layer error of every (slot, layer, head) unit, the relative output error of sparse attention without compensation on 1{,}024 sampled query rows, for each ratio $\eta\in\{1,0.75,0.5,0.25,0\}$ and each density from 0.05 to 0.80 in steps of 0.05.
Alg.~\ref{alg:budget} then solves the schedule by dynamic programming over the work spent.
It uses the errors at $\eta=1$, that is, with the positional table alone, so the schedule does not depend on the ratios.
The density ladder stops at the crossover density $\rho^*$ above which a sparse call is slower than FA3, which caps it at 0.55 on Wan2.1 at 720p. App.~\ref{app:results} further demonstrates this on the kernel roofline analysis.

\begin{algorithm}[t]
\caption{Step-adaptive sparsity budget allocation}
\label{alg:budget}
\begin{algorithmic}[1]
\Require step slots $s=1,\dots,S$, each serving $n_s$ steps; mean amplification $A_s$ over these steps; mean error $e_s(\rho)$ at $\eta=1$; density ladder $\mathcal R=\{0.05,0.10,\dots\}$ up to $\rho^*$; target density $\bar\rho$
\State $F_0(0)\gets 0$;\quad $F_s(b)\gets\infty$ otherwise
  \Comment{$b$: work spent}
\For{$s=1$ to $S$}
  \For{each density $\rho\in\mathcal R$ and work $b$}
    \State $F_s(b)\gets\min\big(F_s(b),\ F_{s-1}(b-n_s\rho)+n_sA_s\,e_s(\rho)\big)$
  \EndFor
\EndFor
\State $b^\star\gets\arg\min_{b\,\le\,\bar\rho\sum_s n_s}F_S(b)$
  \Comment{within fixed budget}
\State $\rho_S,\dots,\rho_1\gets$ backtrack the choices from $F_S(b^\star)$
\Ensure density $\rho_s$ per step slot
\end{algorithmic}
\end{algorithm}

\paragraph{Position--Content Ratio Allocation.}
At the density each slot receives, Alg.~\ref{alg:ratio} allocates the ratios by the 80\% fraction of Sec.~\ref{sec:poco}.
Each head contributes the moves along the lower convex hull of its error over the five ratio levels, and the moves of all heads are applied in decreasing order of error reduction per unit of $\eta_h$ given up, until 80\% of the attainable reduction is reached.
Both solvers run on the CPU, so a new target density needs no further GPU time, and Tab.~\ref{tab:deploy} lists the resulting schedules and mean ratios.

\begin{algorithm}[t]
\caption{Position--content ratio allocation}
\label{alg:ratio}
\begin{algorithmic}[1]
\Require error $e_h(\eta)$ of each step slot, layer, head $h$ at its slot density $\rho_s$, for $\eta\in\mathcal L=\{1,0.75,0.5,0.25,0\}$; fraction $\gamma=0.8$
\For{each step slot, layer, head $h$}
  \State list the moves of $h$: from $\eta=1$, repeatedly step to the lower level $\eta'$ with the largest gain $g=\frac{e_h(\eta)-e_h(\eta')}{\eta-\eta'}$ while $g>0$

\EndFor
\State $\eta_h\gets 1$ for all $h$;\quad $E\gets\operatorname{mean}_h\,e_h(\eta_h)$
\State $E_1\gets E$;\quad $E_{\min}\gets\operatorname{mean}_h\min_{\eta\in\mathcal L}e_h(\eta)$
\For{each listed move $(h,\eta')$ in decreasing gain $g$}
  \State \textbf{if} $E_1-E\ge\gamma\,(E_1-E_{\min})$ \textbf{then break}
  \State $\eta_h\gets\eta'$;\quad update $E$
\EndFor
\Ensure ratio $\eta_h$ per step slot, layer, head
\end{algorithmic}
\end{algorithm}

\paragraph{Sparse Attention Kernel and Fused Operators.}
Before attention, the tokens of each head are packed into its head-adaptive
tile order, so every positional key tile is a contiguous block of 64 keys and
every query tile a block of 128 queries. The sparse kernel is written in
Triton~\cite{triton} for Hopper GPUs. Each program processes one query tile of
one head in a single online-softmax pass over three kinds of keys: its
positional key tiles, loaded whole with TMA; its content keys, gathered by
index with \texttt{cp.async}; and the virtual keys of compensation.
Content keys come from an exact top-$k$ over the pooled-query logits, computed
by a separate kernel before attention, and heads with $\eta_h=1$ skip this
scoring. In an isolated benchmark on Wan2.1-14B at 720p and $\rho=0.25$, a full sparse call takes 74.3\,ms, of which 65.8\,ms is the attention pass and the rest is content selection and compensation, against 161.9\,ms for FA3.
Outside attention, the normalization, RoPE, modulation, and gated-residual
chains of each block are fused into Triton kernels that match the stock
operators bitwise. For Blackwell GPUs, we reimplement the sparse attention
kernel in CuTe DSL on top of the FlashAttention-4~\cite{fa4} forward kernel for
SM100, gathering content keys with the TMA row gather of Blackwell.

\section{Experimental Setup}\label{app:setup}

\paragraph{Evaluation Sets.}
Following SVG-EAR~\cite{svgear}, text-to-video models are evaluated on 50 prompts drawn at random with a fixed seed from the Penguin Benchmark~\cite{hunyuanvideo}.
SVG-EAR uses Penguin prompts rewritten by the VBench team, which are not public, so we rewrite the sampled prompts with the official prompt extension of Wan2.1~\cite{wan}, which runs Qwen2.5-14B-Instruct~\cite{qwen25} and yields a median length of 93 words.
HunyuanVideo and LTX-2.3 use the same 50 extended prompts, with the built-in prompt enhancer of LTX-2.3 disabled. App.~\ref{app:results} also evaluates on the original texts of the same 50 prompts, which have a median length of 22 words, to show that the results do not depend on the prompt extension.
Image-to-video generation uses 50 of the 355 VBench-I2V~\cite{vbenchpp} image--caption pairs, drawn at random, with the official 16:9 crops of the images.
For Wan2.1-I2V, the captions are extended in the same way by Qwen2.5-VL-7B~\cite{qwen25vl} from the image and the caption.
Video editing uses all 73 EditVerseBench~\cite{editverse} cases following the setting of Sol-Attn~\cite{sol}.
All methods generate a given prompt from the same seed, 42 plus the prompt's index in its benchmark, so each video is compared with the dense video from the same initial noise, and the averages are taken over a different noise draw per prompt instead of one fixed seed.

\paragraph{Generation Settings.}
\input{tables/App_Parameter}
Tab.~\ref{tab:gen_settings} summarizes the generation settings.
Wan2.1 samples with UniPC~\cite{unipc} at flow shift 5.0 and classifier-free guidance 5.0, and HunyuanVideo uses the default flow-matching sampler and embedded guidance of its diffusers pipeline.
LTX-2.3 follows the two-stage distilled pipeline of the Sol-Attn setting, which generates 361 frames at $960\times544$ in stage~1 and refines the upsampled 721-frame video at $1920\times1088$ in stage~2.
Following the official warm-up setting of prior work~\cite{svg2,sol,mcsparse} and for a fair comparison, all methods share the warm-up steps listed in Tab.~\ref{tab:gen_settings}, 20\% of the steps or all of stage~1 on LTX-2.3, and the dense first layer. Every dense attention call runs FlashAttention-3~\cite{fa3}, including those inside the baselines.
Each video is generated on a single H100 80 GB GPU, except that Bernini runs all methods on eight H100 GPUs with Ulysses sequence parallelism~\cite{ulysses}.

\input{tables/Baseline_identification}

\paragraph{Baseline Configurations.}
Each baseline runs its official implementation with the configuration of its own scripts. Every baseline computes at least as many query--key pairs as PoCoAttn, and Tab.~\ref{tab:main} reports the real measured densities.
SVG2~\cite{svg2} and SVG-EAR~\cite{svgear} use the query and key cluster counts of their official scripts with top-$p$ values of 0.90 and 0.87. PISA~\cite{pisa} runs at density 0.30 (0.25 on Bernini), and since its released code provides only the kernel, its tokens are ordered in $4\times4\times4$ tiles, which scores 0.3\,dB higher than the raster-order reproduction in Sol-Attn on Wan2.1 at 480p.
Sol-Attn uses one threshold for all steps, $\tau=0.52$ as in MC-Sparse~\cite{mcsparse}, and $\tau=1.18$ for the 15\% row of LTX-2.3.
CalibAtt~\cite{calibatt} has no density parameter at inference, so it runs with its released 720p calibration of nominal density 37.5\%, and its public code supports only Wan2.1 text-to-video. Calibration data under other settings are not provided, so we report its results only on Wan2.1-T2V at 720p.
All baselines and PoCoAttn keep the same dense warm-up described under Generation Settings, for a fair comparison.

As categorized in Tab.~\ref{tab:category}, the baselines differ in whether keys are selected by fixed masks or online at inference time, and in whether the unselected keys are compensated for.
At a matched density, static selection is generally faster than dynamic selection since it has no online selection cost, but less faithful, and methods with compensation are more faithful than those that discard the unselected keys.
PoCoAttn compensates as well, so its comparison with the compensated baselines, SVG-EAR, PISA, and Sol-Attn, is on equal terms, and these include the strongest baseline of every setting.

\paragraph{Metrics.}
We measure fidelity to dense attention with PSNR, SSIM, and LPIPS~\cite{lpips}, and visual quality with VBench~\cite{vbench}.
The fidelity metrics compare each video frame by frame with the dense video of the same prompt and seed at full resolution, and are averaged over frames and then over prompts.
SSIM follows its original definition~\cite{ssim}, with an $11\times11$ Gaussian window of $\sigma=1.5$ on RGB frames, and LPIPS uses AlexNet features with inputs scaled to $[-1,1]$, the default of its official implementation.
Density is the number of query--key pairs a method computes exactly in its sparse attention calls, divided by the number that dense attention computes in the same calls. Following Sol-Attn~\cite{sol}, we report eight VBench dimensions and their mean (AVG).
Unlike the other dimensions, DD labels each video as either dynamic or static and reports the fraction of dynamic videos. With 50 prompts, one video flipping between the two labels changes DD by 2 points and AVG by 0.25, so AVG differences below about 0.5 can arise from DD flips of one or two videos alone.

\paragraph{Efficiency Measurement.}
Latency is measured in separate timing runs with the same generation settings, which discard the first video to exclude compilation and autotuning and report the median over the remaining videos.
DiT time covers the denoising loop including the dense warm-up steps, E2E time adds condition encoding and VAE decoding, and Speedup is the DiT denoising speedup over dense attention.
Attn$\times$ times each sparse attention call as a whole, including selection, clustering, permutation, and compensation, against an FA3 call of the same shape.
All methods on our backend, dense attention included, use our fused operators of App.~\ref{app:impl}, while SVG2 and SVG-EAR keep their official ones.
The lower speedup of SVG2 and SVG-EAR in Tab.~\ref{tab:main} mainly comes from this stronger dense baseline, which is about 1.6$\times$ as fast as the dense run of their official code, where they reach 1.83$\times$ and 1.75$\times$ on Wan2.1-T2V.

\input{tables/pair_wining}

\section{Additional Results}\label{app:results}

\paragraph{Per-Prompt Consistency.}
Fidelity varies widely across prompts. The standard deviation of PSNR over the evaluation set is 4.0--5.9\,dB for every method in Tab.~\ref{tab:main}, comparable to or larger than the gaps between methods, so the means alone do not show whether a gain holds on individual prompts. Since all methods generate the same prompts with the same seeds, we compare them prompt by prompt. Tab.~\ref{tab:paired} reports the paired PSNR difference between PoCoAttn at 25\% density and each baseline with its 95\% confidence interval, together with the number of prompts on which PoCoAttn is higher. Pairing removes the variation shared across methods, and the half-width of the interval is only 0.3--1.0\,dB. PoCoAttn is significantly higher than every baseline in every setting. It wins on 40--50 of the 50 prompts, on 48--50 against the strongest baseline, and on 70 of the 73 editing cases. The lowest counts, 40 and 42, are against SVG-EAR, which also has the largest spread across prompts.

\paragraph{Robustness to Prompt Length.}
Our text-to-video prompts are extended as described in App.~\ref{app:setup}, which raises their median length from 22 to 93 words, whereas PoCoAttn is calibrated on short MovieGenBench~\cite{moviegen} prompts.
Tab.~\ref{tab:prompt_len} evaluates LTX-2.3 on the same 50 prompts before and after extension with one calibration.
PoCoAttn at 25\% leads Sol-Attn at 30\% by 2.23\,dB on the original prompts and 1.86\,dB on the extended ones, and the leads at 15\% are 2.51 and 2.04\,dB. At 25\%, it is more faithful on 50 and 48 of the 50 prompts, so its gain does not come from the extension.
\input{tables/Prompt_length}

\begin{figure}[t]
  \centering
  \includegraphics[width=\columnwidth]{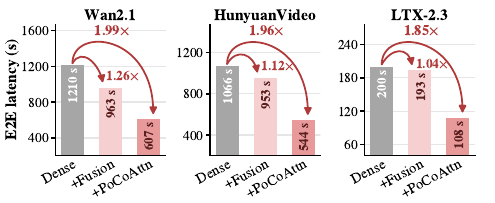}
  \caption{End-to-end latency and speedup across models as acceleration techniques are progressively applied, 15\% density.}
  \label{fig:e2e_latency}
\end{figure}

\paragraph{End-to-End Latency Breakdown.}
Fig.~\ref{fig:e2e_latency} breaks down the E2E latency as the two acceleration techniques are applied in turn. Fusing the non-attention operators alone brings a 1.04--1.26$\times$ speedup over the original dense pipeline, and adding PoCoAttn at 15\% density raises the total to 1.85--1.99$\times$ across Wan2.1, HunyuanVideo, and LTX-2.3. The dense rows of Tab.~\ref{tab:main} already include this fusion, so the speedups reported there isolate the contribution of sparse attention.

\begin{figure}[t]
  \centering
  \includegraphics[width=\columnwidth]{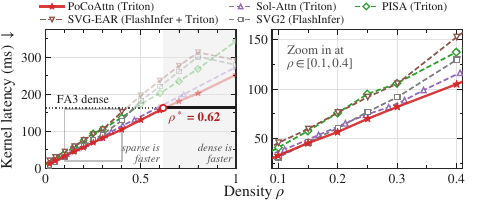}
  \caption{Sparse attention kernel roofline on Wan2.1-T2V-14B, 720p, on the NVIDIA H100.}
  \label{fig:kernel_roofline}
\end{figure}

\paragraph{Sparse Attention Kernel Roofline.}
Fig.~\ref{fig:kernel_roofline} measures the single-call latency of the sparse attention kernels against the executed density $\rho$ on Wan2.1-14B at 720p. Each kernel follows a nearly linear curve whose slope is the cost per computed pair. Our Triton kernel has the smallest slope and crosses FA3 dense attention latest, at $\rho^*=0.62$, whereas the Triton kernels of PISA and Sol-Attn and the FlashInfer backend of SVG2 and SVG-EAR cross at $\rho\approx0.45$--$0.55$ and are 10--50\% slower at $\rho=0.4$. Above the crossover, sparse attention costs more than dense attention while still approximating it, which is why the step-wise allocation of Sec.~\ref{sec:step} caps the density of every slot. With online content scoring included, a full sparse call crosses FA3 at $\rho\approx0.60$, so the density ladder stops at 0.55.

\paragraph{Calibration Cost and Memory Overhead.}
Calibration is a one-time offline cost per generation setting. For Wan2.1-14B at 720p, PoCoAttn requires 44 H100 GPU-hours to calibrate on 8 prompts, and a single calibration supports all density settings and transfers across tasks, as shown in Tab.~\ref{tab:calib_transfer}. In contrast, CalibAtt~\cite{calibatt} requires recalibration for each density and task setting, taking 89.6 H100 GPU-hours with its default 64 prompts. The cost can be reduced further, since the calibration is insensitive to the number of prompts. Fig.~S4 varies it from 1 to 16 on LTX-2.3 under a uniform schedule: PSNR stays within 0.12\,dB and SSIM and LPIPS within 0.002 across the range, and even a single prompt matches our default of 8. A single prompt also yields the same step schedule as 8, and with it the whole calibration takes about 7 H100 GPU-hours. The positional prior is thus a property of the model rather than of the calibration data.
At inference, PoCoAttn adds little memory: its tables store a per-head ranking of key tiles, only those of the current step slot reside on the GPU while the rest stay in host memory, and the workspaces of a call are allocated once and reused by every call. As Tab.~\ref{tab:memory} shows, it adds at most 2.4\,GB to the peak memory of dense attention, comparable to the dynamic baselines, such as the 2.2\,GB that Sol-Attn~\cite{sol} adds on LTX-2.3. CalibAtt, which is also calibrated offline, adds 9.6\,GB on Wan2.1-T2V, four times as much, because it keeps a compute list for every step, layer, head, and query tile on the GPU, whereas each table of PoCoAttn serves all steps of its slot.

\input{tables/Mem_overhead}

\paragraph{Calibration Table Transfer.}

PoCoAttn calibrates each model once under its evaluation setting, and Tab.~\ref{tab:calib_transfer} tests whether the tables carry over to a new resolution or task without recalibration.
For resolution, the per-head offset priors measured at 480p are rasterized onto the 720p token grid, with offsets beyond the 480p range clamped to the nearest measured offset, and the box shapes are reselected on this grid, while the ratios and step schedule are copied from 480p, so no new calibration run is needed.
At 25\% density, the transferred tables keep 75\% of the keys selected by the native 720p tables on sampled layers, against 25\% for random selection.
On Wan2.1-T2V at 720p, they reach 27.52\,dB, 0.28\,dB below native calibration with nearly the same speedup, and remain 3.1\,dB above the strongest baseline in Tab.~\ref{tab:main}. For task, the 720p tables calibrated on Wan2.1-T2V are applied unchanged to Wan2.1-I2V, which uses the same token grid.
The two models are separate checkpoints, and the I2V model additionally conditions on the input image, yet their independently calibrated tables keep 80\% of the same keys at 25\% density.
On the 50 image--prompt pairs, the T2V tables are on par with the native I2V tables at nearly the same speedup, reaching 28.56 against 28.38\,dB, with wins on 30 of 50 pairs, and remain 2.2\,dB above the strongest baseline in Tab.~\ref{tab:main}.
The calibrated tables therefore transfer across resolution and task without recalibration.

\input{tables/Transfer}
\begin{table}[b]
\centering

\caption{\textbf{Full component ablation} on Wan2.1-T2V-14B, 720p on a \textbf{20-prompt subset} of the evaluation set. 25\% density. Pos.: positional keys from the table; Cont.: online content keys; Comp.: compensation.}
\vspace{-0.3em}
\label{tab:abl_loo}
\footnotesize
\setlength{\tabcolsep}{2pt}
\resizebox{\columnwidth}{!}{%
\begin{tabular}{ccc|ccccc}
\toprule
\textbf{Pos.} & \textbf{Cont.} & \textbf{Comp.}
& \textbf{PSNR}$\uparrow$
& \textbf{SSIM}$\uparrow$
& \textbf{LPIPS}$\downarrow$
& \textbf{Attn}$\times$ $\uparrow$
& \textbf{Speedup}$\uparrow$ \\
\midrule
\cmark &        &        & 24.17 & 0.814 & 0.147 & \textbf{2.38$\times$} & \textbf{1.46$\times$} \\
       & \cmark &        & 25.58 & 0.822 & 0.130 & 2.24$\times$ & 1.42$\times$ \\
\cmark & \cmark &        & 26.12 & 0.850 & 0.111 & 2.27$\times$ & 1.43$\times$ \\
\cmark &        & \cmark & 26.07 & 0.858 & 0.109 & 2.18$\times$ & 1.43$\times$ \\
       & \cmark & \cmark & 27.12 & 0.864 & 0.094 & 1.87$\times$ & 1.36$\times$ \\
\rowcolor{oursred}
\cmark & \cmark & \cmark & \textbf{27.73} & \textbf{0.885} & \textbf{0.082} & 2.07$\times$ & 1.40$\times$ \\
\bottomrule
\end{tabular}%
}
\vspace{-1em}
\end{table}

\paragraph{Leave-One-Out Ablation.}

To further support the component ablation in Sec.~\ref{sec:ablation}, we evaluate all combinations of the three components, which adds the two variants with compensation that complete the leave-one-out comparison.
Tab.~\ref{tab:abl_loo} reports the results on a 20-prompt subset of the evaluation set, under the same setting as Tab.~\ref{tab:abl_component}.
The subset tracks the full set closely: the four rows shared with Tab.~\ref{tab:abl_component} differ by at most 0.3~dB, with the full method at 27.73~dB on the subset and 27.81~dB on all 50 prompts.
Removing any single component lowers fidelity, by 0.6~dB without positional keys and by about 1.6~dB without content selection or without compensation, so the three components are complementary rather than redundant.
The positional-content collaboration improves fidelity and speed at the same time: compared with selecting all keys online, it raises PSNR by 0.6~dB and the attention speedup from 1.87$\times$ to 2.07$\times$, since the positional keys come from the table at no online cost.
Compensation then adds 1.6~dB and reduces LPIPS from 0.111 to 0.082 while lowering the denoising speedup by only 0.03$\times$.
These results confirm that PoCoAttn is effective because its components work together, each recovering an error the other two leave behind.

\section{More Qualitative Results}\label{app:qual}
\begin{figure*}[t]
\centering
\includegraphics{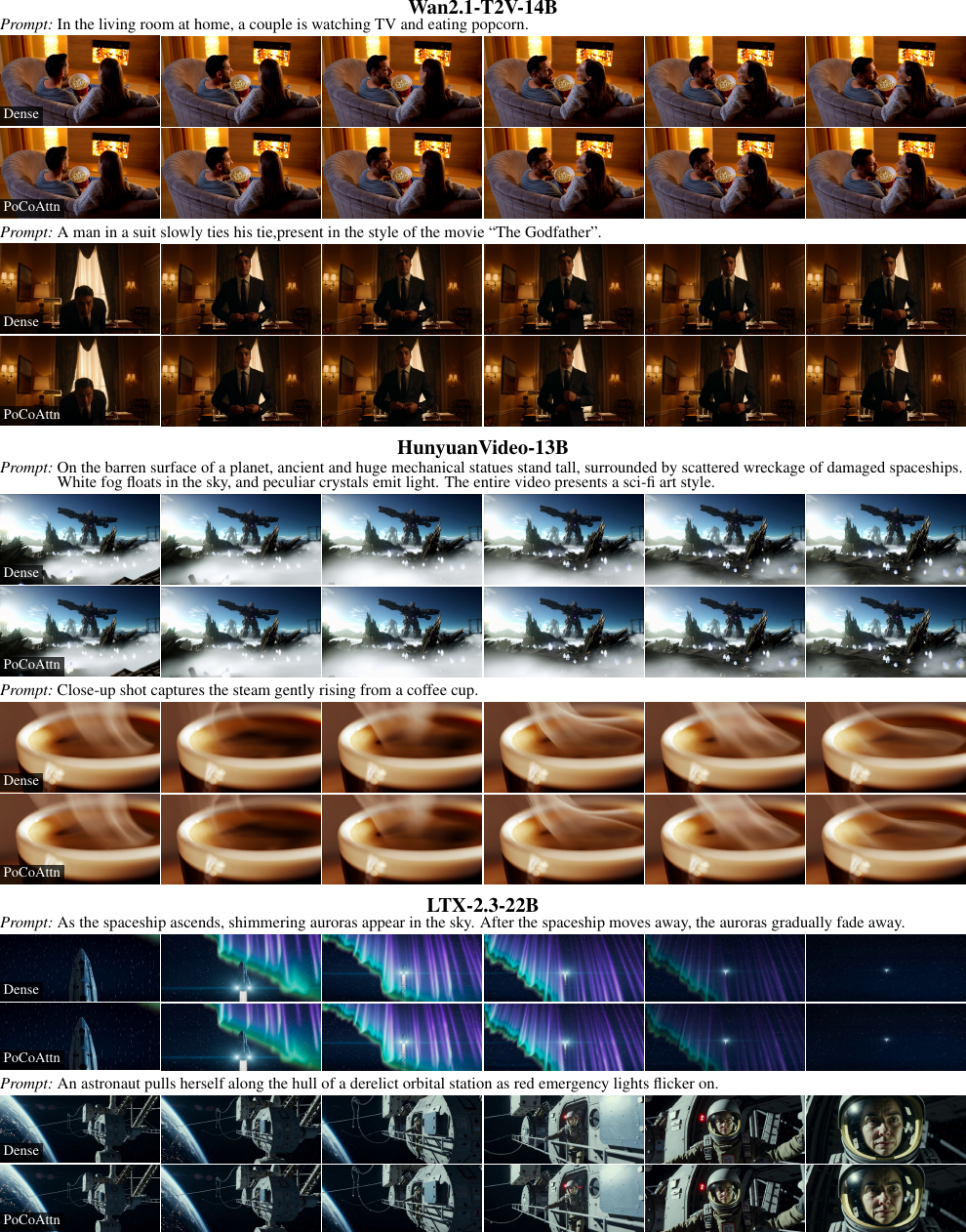}
\caption{\textbf{Qualitative comparison on text-to-video generation.} Frames generated with dense attention (FA3) and with PoCoAttn at 25\% density from the same prompt and seed. Following Sol-Attn~\cite{sol}, LTX-2.3 rows use its high-quality generation setting.}
\label{fig:qual_t2v}
\end{figure*}

\begin{figure*}[t]
\centering
\includegraphics{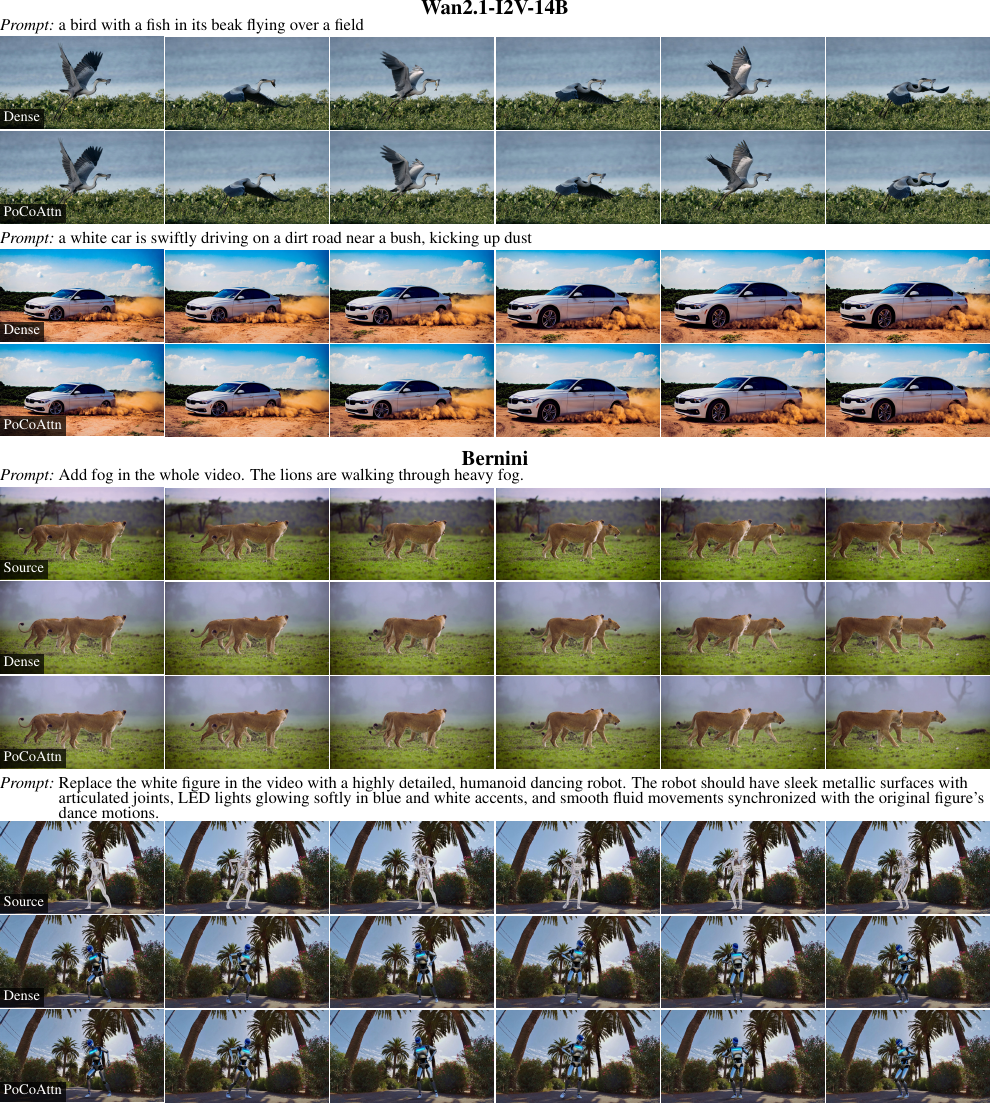}
\caption{\textbf{Qualitative comparison on image-to-video generation and video editing.} Frames generated with dense attention (FA3) and with PoCoAttn at 25\% density from the same input and seed; for video editing, the first row is the source video.}
\label{fig:qual_i2v_v2v}
\end{figure*}

\begin{figure*}[t]
\centering
\includegraphics[width=\textwidth]{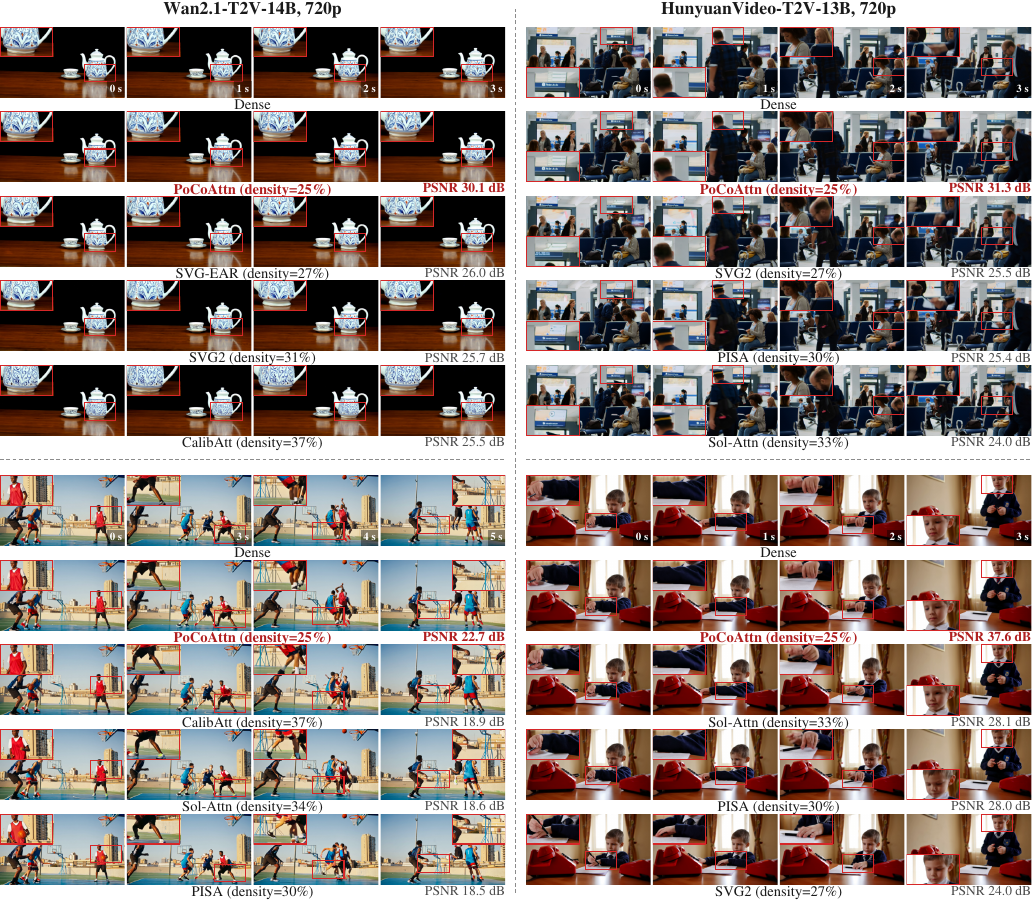}
\caption{\textbf{Detailed comparison with baselines} on Wan2.1-T2V-14B and HunyuanVideo-13B at 720p. Each block shows dense attention, PoCoAttn at 25\% density, and the three baselines with the highest PSNR on that prompt, with PSNR against the dense video. Red boxes mark the magnified regions.}
\label{fig:qual_detail}
\end{figure*}

Fig.~\ref{fig:qual_t2v} and Fig.~\ref{fig:qual_i2v_v2v} \textbf{on the following pages} show PoCoAttn at 25\% density next to dense attention on every evaluated model and task, where it closely reproduces the dense videos.
Fig.~\ref{fig:qual_detail} further compares PoCoAttn with the three strongest baselines on each prompt, where the baselines alter object details and human motion that PoCoAttn preserves.

%% file: tables/Settings.tex
\begin{table*}[t]
\centering
\small
\setlength{\tabcolsep}{8pt}
\caption{\textbf{Deployment details per model}. Schedules list our allocated budget of the step slots from early to late and the mean positional ratio $\eta$ solved by our method.}
\label{tab:deploy}
\begin{tabular}{@{}ll|lclc@{}}
\toprule
\multirow{2}{*}{\textbf{Model}} & \multirow{2}{*}{\textbf{Step Slots}} & \multicolumn{2}{|c}{\textbf{25\% density}} & \multicolumn{2}{c}{\textbf{15\% density}} \\
\cmidrule(lr){3-4} \cmidrule(lr){5-6}
& & \textbf{Budget Schedule} & \textbf{\boldmath Mean $\eta$} & \textbf{Budget Schedule} & \textbf{\boldmath Mean $\eta$} \\
\midrule
Wan2.1-T2V-14B & $10,15,\dots,45$ & .55 .55 .50 .25 .15 .10 .05 .05 & 0.868 & .55 .35 .20 .10 .05 .05 .05 .05 & 0.864 \\
Wan2.1-I2V-14B & $10,15,\dots,45$ & .55 .55 .45 .30 .15 .10 .05 .05 & 0.895 & .55 .35 .20 .10 .05 .05 .05 .05 & 0.894 \\
HunyuanVideo-13B & $10,15,\dots,45$ & .55 .55 .40 .25 .20 .10 .10 .05 & 0.886 & .45 .30 .20 .15 .10 .05 .05 .05 & 0.876 \\
LTX-2.3-22B & $1,2,3$ of stage~2 & .40 .20 .15 & 0.794 & .25 .10 .10 & 0.788 \\
\bottomrule
\end{tabular}
\end{table*}

%% file: tables/App_Parameter.tex
\begin{table}[t]
\centering
\caption{\textbf{Generation and warm-up settings.} LTX-2.3 runs 8 steps in stage~1 and 3 steps in stage~2 following Sol-Attn~\cite{sol}.}
\label{tab:gen_settings}
\footnotesize
\setlength{\tabcolsep}{3pt}
\resizebox{\columnwidth}{!}{%
\begin{tabular}{@{}lcccc@{}}
\toprule
Model & Resolution & Frames & Steps & Warm-up steps \\
\midrule
Wan2.1-T2V-14B & $1280\times720$ & 81 & 50 & 10 \\
Wan2.1-I2V-14B & $1280\times720$ & 81 & 50 & 10 \\
HunyuanVideo-13B & $1280\times720$ & 129 & 50 & 10 \\
LTX-2.3-22B & $1920\times1088$ & 721 & $8+3$ & 8 (stage 1) \\
Bernini & $1280\times720$ & 81 & 40 & 8 \\
\bottomrule
\end{tabular}}
\end{table}

%% file: tables/Baseline_identification.tex
\begin{table}[t]
\centering
\caption{\textbf{Categorization of the sparse attention baselines used in our experiments.}
\emph{Static}/\emph{Dynamic}: keys are selected by fixed masks / by online selection at inference time.
\emph{Offline calib.}: the sparse pattern is calibrated offline before inference.
\emph{Comp.}: the contribution of unselected keys is compensated rather than discarded.
PoCoAttn combines all four.}
\label{tab:category}
\footnotesize
\setlength{\tabcolsep}{8.5pt}
\begin{tabular}{lcccc}
\toprule
& \multicolumn{2}{c}{\textbf{Key selection}} & \textbf{Offline} & \\
\cmidrule(lr){2-3}
\textbf{Method} & \textbf{Static} & \textbf{Dynamic} & \textbf{calib.} & \textbf{Comp.} \\
\midrule
RadialAttn~\cite{radial}  & \tabyes & \tabno  & \tabno  & \tabno  \\
CalibAtt~\cite{calibatt}  & \tabyes & \tabno  & \tabyes & \tabno  \\
SVG2~\cite{svg2}          & \tabno  & \tabyes & \tabno  & \tabno  \\
SVG-EAR~\cite{svgear}     & \tabno  & \tabyes & \tabno  & \tabyes \\
PISA~\cite{pisa}          & \tabno  & \tabyes & \tabno  & \tabyes \\
Sol-Attn~\cite{sol}       & \tabno  & \tabyes & \tabno  & \tabyes \\
\rowcolor{oursred}
\textbf{PoCoAttn}         & \tabyes & \tabyes & \tabyes & \tabyes \\
\bottomrule
\end{tabular}
\end{table}

%% file: tables/pair_wining.tex
\begin{table}[t]
\centering
\caption{\textbf{Paired per-prompt comparison} of PoCoAttn at 25\% density against every baseline in Tab.~\ref{tab:main} and Tab.~\ref{tab:bernini} that runs at the same or a higher density than PoCoAttn (25.0--36.6\%), over the 50 prompts or image--prompt pairs of each generation setting and the 73 editing cases of Bernini. $\Delta$PSNR is the mean paired difference $\pm$ the half-width of its 95\% confidence interval, and Wins counts the cases on which PoCoAttn has the higher PSNR. $^\ddagger$ marks the strongest baseline of each setting.}
\label{tab:paired}
\setlength{\tabcolsep}{22pt}
\resizebox{\columnwidth}{!}{%
\begin{tabular}{l|cc}
\toprule
\textbf{Baseline} & \textbf{$\Delta$PSNR (dB)} & \textbf{Wins} \\
\midrule
\multicolumn{3}{c}{\textbf{Wan2.1-T2V-14B} \textit{(Text-to-Video, 720p)}} \\
\midrule
RadialAttn~\cite{radial} & $+5.31\pm0.42$ & 50/50 \\
SVG2~\cite{svg2} & $+4.21\pm0.82$ & 47/50 \\
SVG-EAR~\cite{svgear} & $+3.52\pm0.99$ & 40/50 \\
PISA~\cite{pisa}$^\ddagger$ & $+3.38\pm0.39$ & 50/50 \\
Sol-Attn~\cite{sol} & $+3.68\pm0.54$ & 49/50 \\
CalibAtt~\cite{calibatt} & $+3.60\pm0.48$ & 50/50 \\
\midrule
\multicolumn{3}{c}{\textbf{Wan2.1-I2V-14B} \textit{(Image-to-Video, 720p)}} \\
\midrule
SVG2~\cite{svg2} & $+3.99\pm0.72$ & 49/50 \\
SVG-EAR~\cite{svgear} & $+2.87\pm0.87$ & 42/50 \\
PISA~\cite{pisa}$^\ddagger$ & $+2.01\pm0.38$ & 48/50 \\
Sol-Attn~\cite{sol} & $+3.27\pm0.73$ & 47/50 \\
\midrule
\multicolumn{3}{c}{\textbf{HunyuanVideo} \textit{(Text-to-Video, 720p)}} \\
\midrule
SVG2~\cite{svg2} & $+5.48\pm0.99$ & 50/50 \\
PISA~\cite{pisa}$^\ddagger$ & $+5.00\pm0.63$ & 50/50 \\
Sol-Attn~\cite{sol} & $+6.34\pm0.47$ & 50/50 \\
\midrule
\multicolumn{3}{c}{\textbf{LTX-2.3} \textit{(distilled, Text-to-Video, 1080p)}} \\
\midrule
RadialAttn~\cite{radial} & $+9.42 \pm 0.81$ & 50/50 \\
PISA~\cite{pisa} & $+2.81 \pm 0.27$ & 49/50 \\
Sol-Attn~\cite{sol} (30\%)$^\ddagger$ & $+1.86 \pm 0.26$ & 48/50 \\
\midrule
\multicolumn{3}{c}{\textbf{Bernini} \textit{(Video Editing, 720p)}} \\
\midrule
PISA~\cite{pisa}$^\ddagger$ & $+3.87\pm0.51$ & 70/73 \\
Sol-Attn~\cite{sol} & $+4.25\pm0.44$ & 72/73 \\
\bottomrule
\end{tabular}}
\end{table}

%% file: tables/Prompt_length.tex
\begin{table}[t]
\centering
\caption{\textbf{Robustness to prompt length} on LTX-2.3, on the same 50 prompts in their original and extended forms. Speedups on the extended prompts are those of Tab.~\ref{tab:main}; those on the original prompts are timed on these prompts with the same protocol.}
\vspace{-3mm}
\label{tab:prompt_len}
\footnotesize
\setlength{\tabcolsep}{3pt}
\renewcommand{\arraystretch}{0.92}
\resizebox{\columnwidth}{!}{%
\begin{tabular}{l|cccccc}
\toprule
\textbf{Method}
& \textbf{PSNR}$\uparrow$
& \textbf{SSIM}$\uparrow$
& \textbf{LPIPS}$\downarrow$
& \textbf{Density}$\downarrow$
& \textbf{\boldmath Attn$\times$}$\uparrow$
& \textbf{Speedup}$\uparrow$ \\
\midrule
\multicolumn{7}{c}{\textbf{Original prompts}} \\
\midrule
Dense (\textbf{FA3}~\cite{fa3}) & -- & -- & -- & 100\% & 1.00$\times$ & 1.00$\times$ \\
PISA~\cite{pisa} & 30.37 & 0.868 & 0.116 & 30.0\% & 1.28$\times$ & 1.18$\times$ \\
Sol-Attn~\cite{sol} (30\%) & 31.41 & 0.882 & 0.102 & 31.3\% & 1.40$\times$ & 1.25$\times$ \\
Sol-Attn~\cite{sol} (15\%) & 28.86 & 0.847 & 0.145 & 15.5\% & 2.95$\times$ & 1.86$\times$ \\
\rowcolor{oursred}
\textbf{PoCoAttn} (25\%) & \textbf{33.64} & \textbf{0.917} & \textbf{0.073} & 25.0\% & 2.27$\times$ & 1.63$\times$ \\
\rowcolor{oursred}
\textbf{PoCoAttn} (15\%) & 31.37 & 0.892 & 0.101 & 15.0\% & \textbf{3.39$\times$} & \textbf{1.96$\times$} \\
\midrule
\multicolumn{7}{c}{\textbf{Extended prompts} (used in Tab.~\ref{tab:main})} \\
\midrule
Dense (\textbf{FA3}~\cite{fa3}) & -- & -- & -- & 100\% & 1.00$\times$ & 1.00$\times$ \\
PISA~\cite{pisa} & 31.14 & 0.880 & 0.123 & 30.0\% & 1.27$\times$ & 1.18$\times$ \\
Sol-Attn~\cite{sol} (30\%) & 32.10 & 0.892 & 0.108 & 31.3\% & 1.41$\times$ & 1.26$\times$ \\
Sol-Attn~\cite{sol} (15\%) & 29.66 & 0.861 & 0.152 & 15.4\% & 2.99$\times$ & 1.87$\times$ \\
\rowcolor{oursred}
\textbf{PoCoAttn} (25\%) & \textbf{33.95} & \textbf{0.917} & \textbf{0.081} & 25.0\% & 2.27$\times$ & 1.65$\times$ \\
\rowcolor{oursred}
\textbf{PoCoAttn} (15\%) & 31.69 & 0.893 & 0.111 & 15.0\% & \textbf{3.48$\times$} & \textbf{2.02$\times$} \\
\bottomrule
\end{tabular}%
}
\vspace{-1.5em}
\end{table}

%% file: tables/Mem_overhead.tex
\begin{table}[t]
\centering
\caption{\textbf{Peak GPU memory} (GB) allocated during generation.}
\label{tab:memory}
\footnotesize
\setlength{\tabcolsep}{3.5pt}
\begin{tabular}{l*{2}{w{c}{1.1cm}}w{c}{1.75cm}w{c}{1.25cm}}
\toprule
\multirow{2}{*}{\textbf{Method}} & \multicolumn{2}{c}{\textbf{Wan2.1}} & \textbf{HunyuanVideo} & \textbf{LTX-2.3} \\
\cmidrule(lr){2-3}\cmidrule(lr){4-4}\cmidrule(lr){5-5}
 & T2V & I2V & T2V & T2V \\
\midrule
Dense    & 58.2 & 64.6 & 51.0 & 75.3 \\
SVG2~\cite{svg2}          & 58.7 & 65.1 & 58.4 & --   \\
SVG-EAR~\cite{svgear}     & 58.7 & 65.1 & --   & --   \\
PISA~\cite{pisa}          & 59.3 & 65.8 & 51.0 & 75.3 \\
Sol-Attn~\cite{sol}       & 58.2 & 64.6 & 51.8 & 77.5 \\
CalibAtt~\cite{calibatt}  & 67.8 & --   & --   & --   \\
\rowcolor{oursred}
\textbf{PoCoAttn} & 60.5 & 66.0 & 53.4 & 76.1 \\
\bottomrule
\end{tabular}
\end{table}

%% file: tables/Transfer.tex
\providecolor{oursred}{RGB}{255,236,236}
\begin{table}[t]
\centering
\caption{\textbf{Calibration table transfer} on Wan2.1 at 720p, over the 50 evaluation prompts or image--prompt pairs, at 25\% density and nearly the same speedup as the native tables (1.39--1.40$\times$). Overlap is the share of keys that the transferred tables select in common with the native ones, against 25\% for random selection. $\Delta$PSNR is the paired difference from the native tables $\pm$ the half-width of its 95\% confidence interval.}
\label{tab:calib_transfer}
\small
\setlength{\tabcolsep}{4pt}
\resizebox{\columnwidth}{!}{%
\begin{tabular}{@{}lccccc@{}}
\toprule
\textbf{Tables} & \textbf{Overlap} & \textbf{\boldmath PSNR$\uparrow$} & \textbf{\boldmath SSIM$\uparrow$} & \textbf{\boldmath LPIPS$\downarrow$} & \textbf{\boldmath $\Delta$PSNR} \\
\midrule
\multicolumn{6}{c}{\textbf{Cross-resolution transfer} (480p$\rightarrow$720p)} \\
\midrule
\rowcolor{oursred}
Native & -- & 27.81 & 0.884 & 0.081 & -- \\
From 480p & 75\% & 27.52 & 0.879 & 0.085 & \textbf{\boldmath$-$0.28}\,{\scriptsize$\pm$0.18} \\
\midrule
\multicolumn{6}{c}{\textbf{Cross-task transfer} (T2V$\rightarrow$I2V)} \\
\midrule
\rowcolor{oursred}
Native & -- & 28.38 & 0.883 & 0.067 & -- \\
From T2V & 80\% & 28.56 & 0.887 & 0.063 & \textbf{\boldmath$+$0.18}\,{\scriptsize$\pm$0.16} \\
\bottomrule
\end{tabular}}
\end{table}